\documentclass{article}

\PassOptionsToPackage{numbers, compress}{natbib}

\usepackage[preprint]{neurips_2024}

\usepackage[utf8]{inputenc} 
\usepackage[T1]{fontenc}    
\usepackage{hyperref}       
\usepackage{url}            
\usepackage{booktabs}       
\usepackage{amsfonts}       
\usepackage{nicefrac}       
\usepackage{microtype}      
\usepackage{xcolor}         
\usepackage{graphicx}
\usepackage{subfig}
\usepackage{float}

\def\name{Vidu}

\title{\name{}: a Highly Consistent, Dynamic and Skilled Text-to-Video Generator with Diffusion Models}

\author{%
  Fan Bao$^{\dagger 1,2}$ \quad Chendong Xiang\thanks{Second authors listed alphabetically. $^\ddagger$The corresponding author.}~~$^{1,2}$ \quad Gang Yue$^{*2}$ \quad Guande He$^{*1,2}$ \quad \textbf{Hongzhou Zhu}$^{*1,2}$ \\
  \textbf{Kaiwen Zheng}$^{*1,2}$ \quad \textbf{Min Zhao}$^{*1,2}$ \quad \textbf{Shilong Liu}$^{*1,2}$ \quad \textbf{Yaole Wang}$^{*2}$ \quad \textbf{Jun Zhu}$^{\ddagger 1,2}$\\
$^\dagger$ bf19@mails.tsinghua.edu.cn; \quad $^\ddagger$  dcszj@tsinghua.edu.cn
  \\
  $^1$Tsinghua University \quad \quad $^2$ShengShu \\
  \url{https://www.shengshu-ai.com/vidu}
}

\begin{document}

\maketitle

\begin{abstract}
  We introduce \name{}, a high-performance text-to-video generator that is capable of producing 1080p videos up to 16 seconds in a single generation.
  \name{} is a diffusion model with U-ViT as its backbone, which unlocks the scalability and the capability for handling long videos.
  \name{} exhibits strong coherence and dynamism, and is capable of generating both realistic and imaginative videos, as well as understanding some professional photography techniques, on par with Sora -- the most powerful reported text-to-video generator. 
  Finally, we perform initial experiments on other controllable video generation, including canny-to-video generation, video prediction and subject-driven generation, which demonstrate promising results.
\end{abstract}

\section{Introduction}

Diffusion models have obtained breakthrough progress on generating high-quality images, videos and other types of data, outperforming alternative approaches like auto-regressive networks. Previously, video generation models primarily relied on diffusion models~\cite{sohl2015deep,ho2020denoising,song2020score} with the U-Net backbone~\cite{ronneberger2015u}, and focused on a single limited duration like 4 seconds~\cite{ho2022imagen,blattmann2023align,girdhar2023emu,blattmann2023stable}.
Our model, \name{}, demonstrates that a text-to-video diffusion model with U-ViT~\cite{bao2023all,bao2023one} as its backbone can break this duration limitation by leveraging the scalability and the long sequence modeling ability of a transformer~\cite{vaswani2017attention}. \name{} is capable of producing 1080p videos up to 16 seconds in a single generation, as well as images as videos of a single frame.

Additionally, \name{} exhibits strong coherence and dynamism, and is capable of generating both realistic and imaginative videos. 
\name{} also has a preliminary understanding of some professional photography techniques, such as transitions, camera movements, lighting effects and emotional portrayal.
We observe that to some extent, the generation performance of \name{} is comparable with that of Sora~\cite{videoworldsimulators2024}, which is currently the most powerful text-to-video generator, much better than the other text-to-video generators.
Finally, we perform initial experiments on other controllable video generation, including canny-to-video generation~\cite{zhang2023adding}, video prediction and subject-driven generation~\cite{ruiz2023dreambooth}. All of them demonstrate promising results.

\section{Text-to-Video Generation}

\begin{figure}[t]
    \centering
    \includegraphics[width=.9\linewidth]{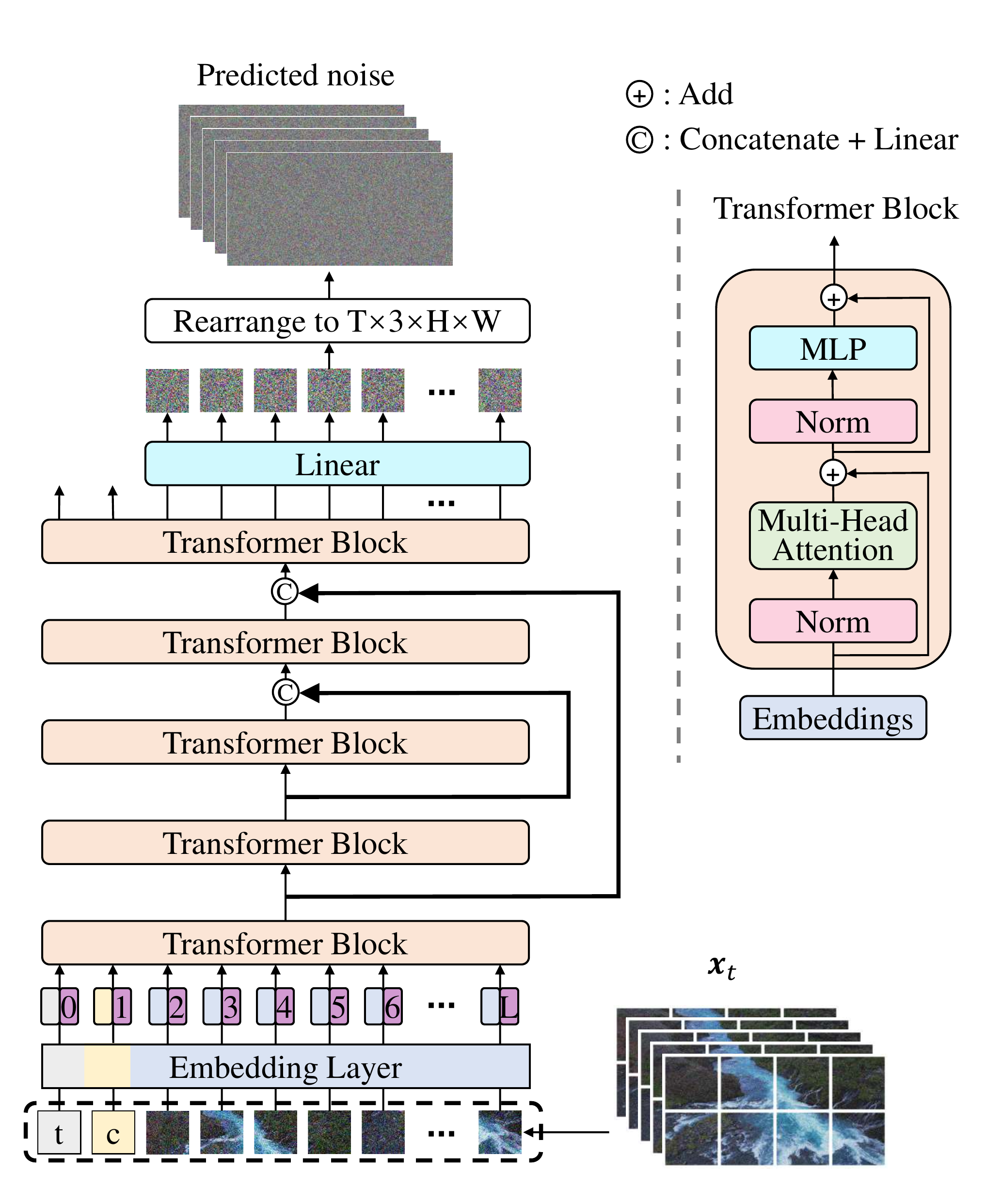}
    \caption{The U-ViT architecture for predicting the noise in videos.}
    \label{fig:uvit}
\end{figure}

\name{} firstly employs a video autoencoder~\cite{kingma2013auto} to reduce both the spatial and temporal dimensions of videos for efficient training and inference. After that, \name{} employs a U-ViT~\cite{bao2023all} as the noise prediction network to model these compressed representations. Specifically, as shown in Figure~\ref{fig:uvit}, U-ViT splits the compressed videos into 3D patches, treats all inputs including the time, text condition and noisy 3D patches as tokens, and employs long skip connections between shallow and deep layers in a transformer. By leveraging the ability of transformers to process variable-length sequences, \name{} can handle videos with variable durations.

\name{} is trained on vast amount of text-video pairs, and it is infeasible to have all videos labeled by humans. To address it, we firstly train a high-performance video captioner optimized for understanding dynamic information in videos, and then automatically annotate all the training videos using this captioner. During inference, we apply the re-captioning technique~\cite{betker2023improving} to rephrase user inputs into a form that is more suitable for the model.

\clearpage
\subsection{Generating Videos of Different Lengths}

\begin{figure}[H]
    \centering
    \subfloat[16 seconds. Prompt: A person clad in a space suit with a helmet and equipped with a chest light and arm device is seen closely examining and interacting with a variety of plants in a lush, indoor botanical setting.]{\includegraphics[width=\linewidth]{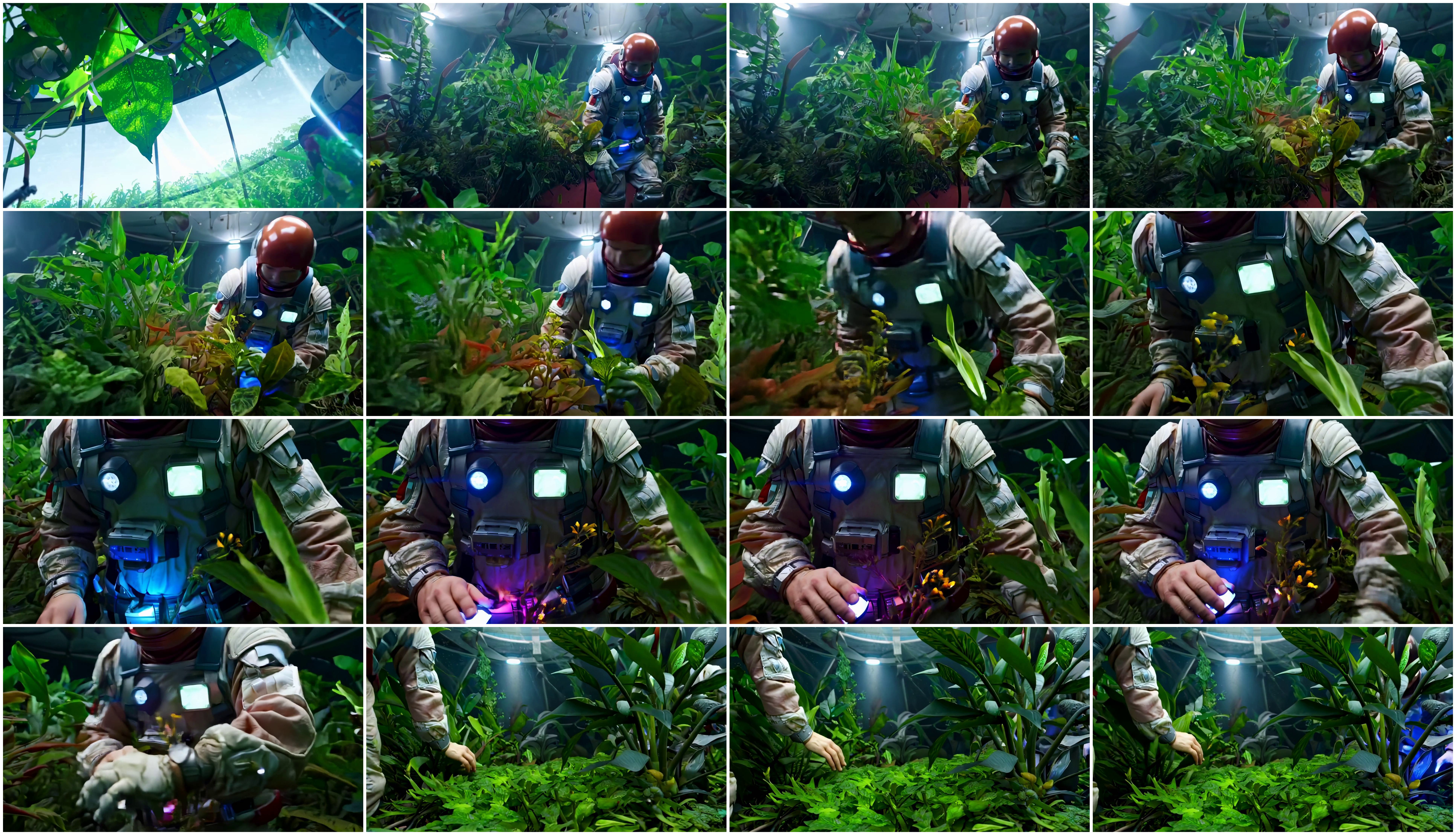}} \\
    \subfloat[8 seconds. Prompt: A desolate lunar landscape with craters and a large moon in the sky transitions to a warmly lit interior of a spacecraft-like structure where a group of people are engaged in various activities.]{\includegraphics[width=\linewidth]{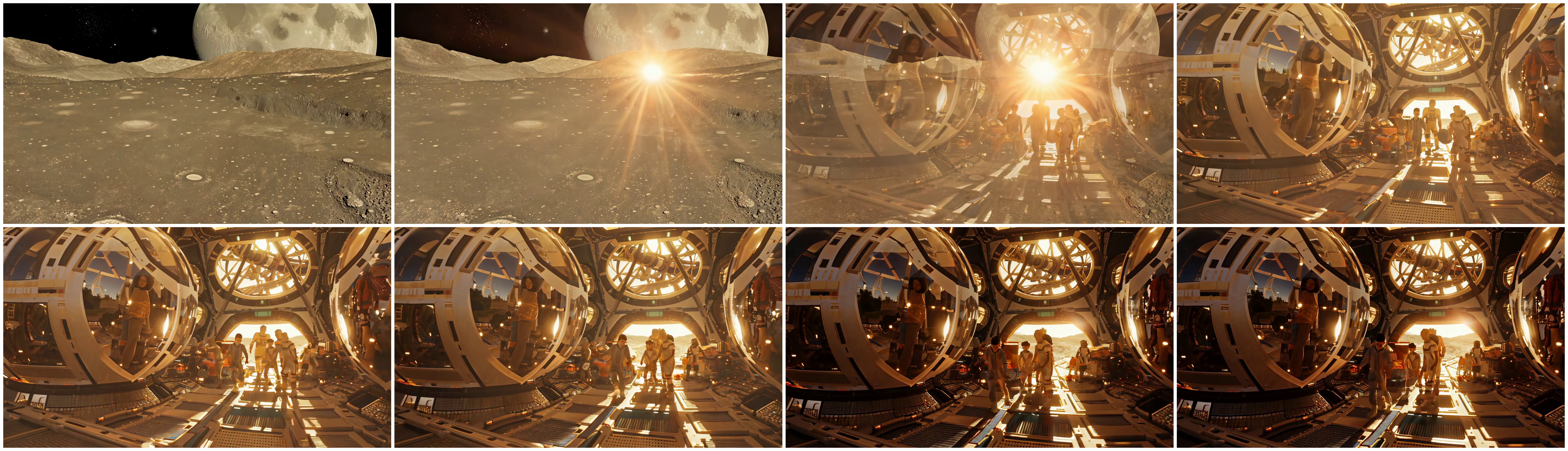}} \\
    \subfloat[Image. Prompt: An exquisite silverware piece, aesthetically adorned with intricate patterns and scenes, exhibits the detailed artisanship and metallic sheen.]{\includegraphics[width=0.49\linewidth]{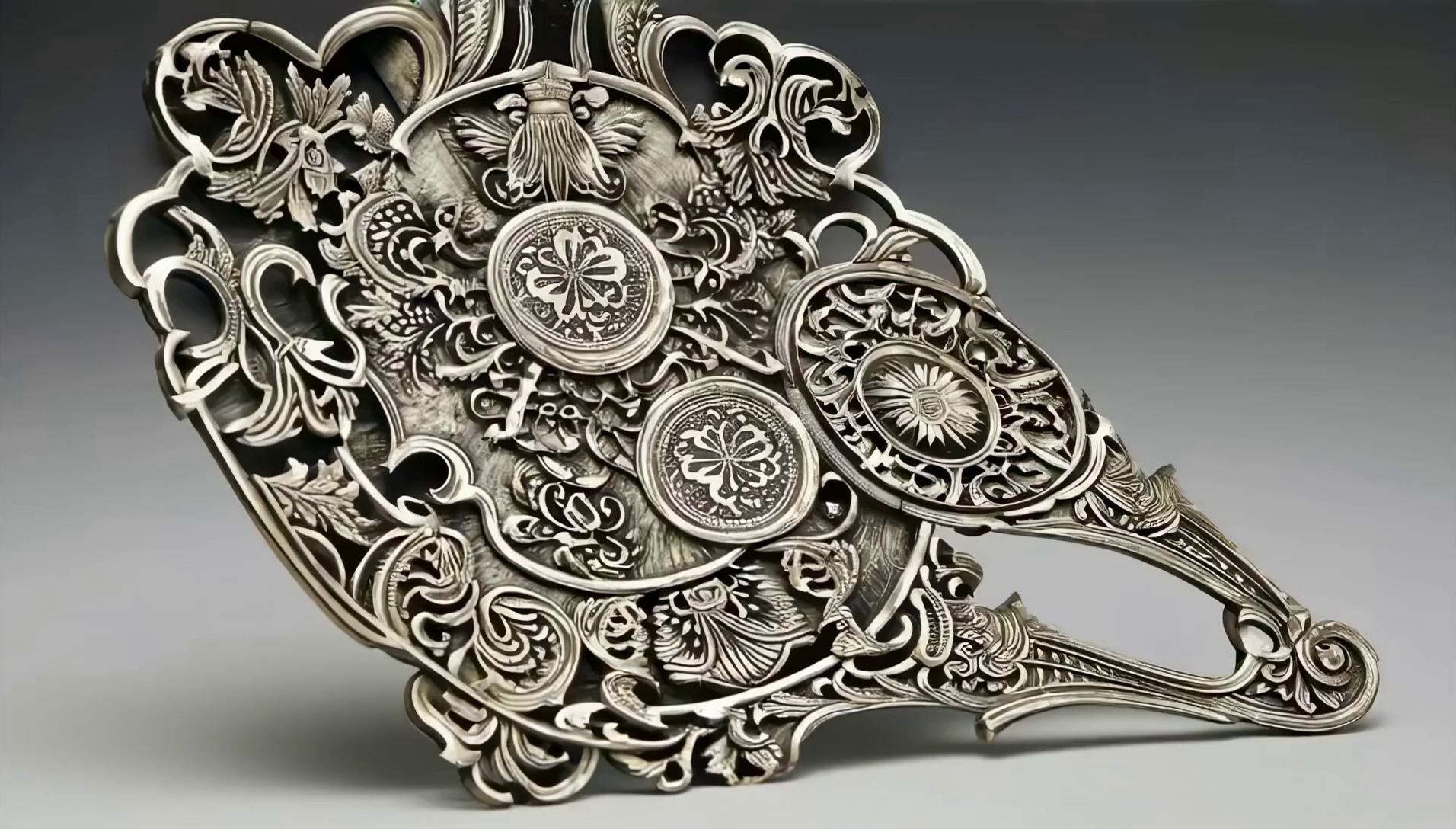}} \hfill
    \subfloat[Image. Prompt: Under the veil of nightfall, a rose reveals its subtle, exquisite beauty in the gentle moonlight.]{\includegraphics[width=0.49\linewidth]{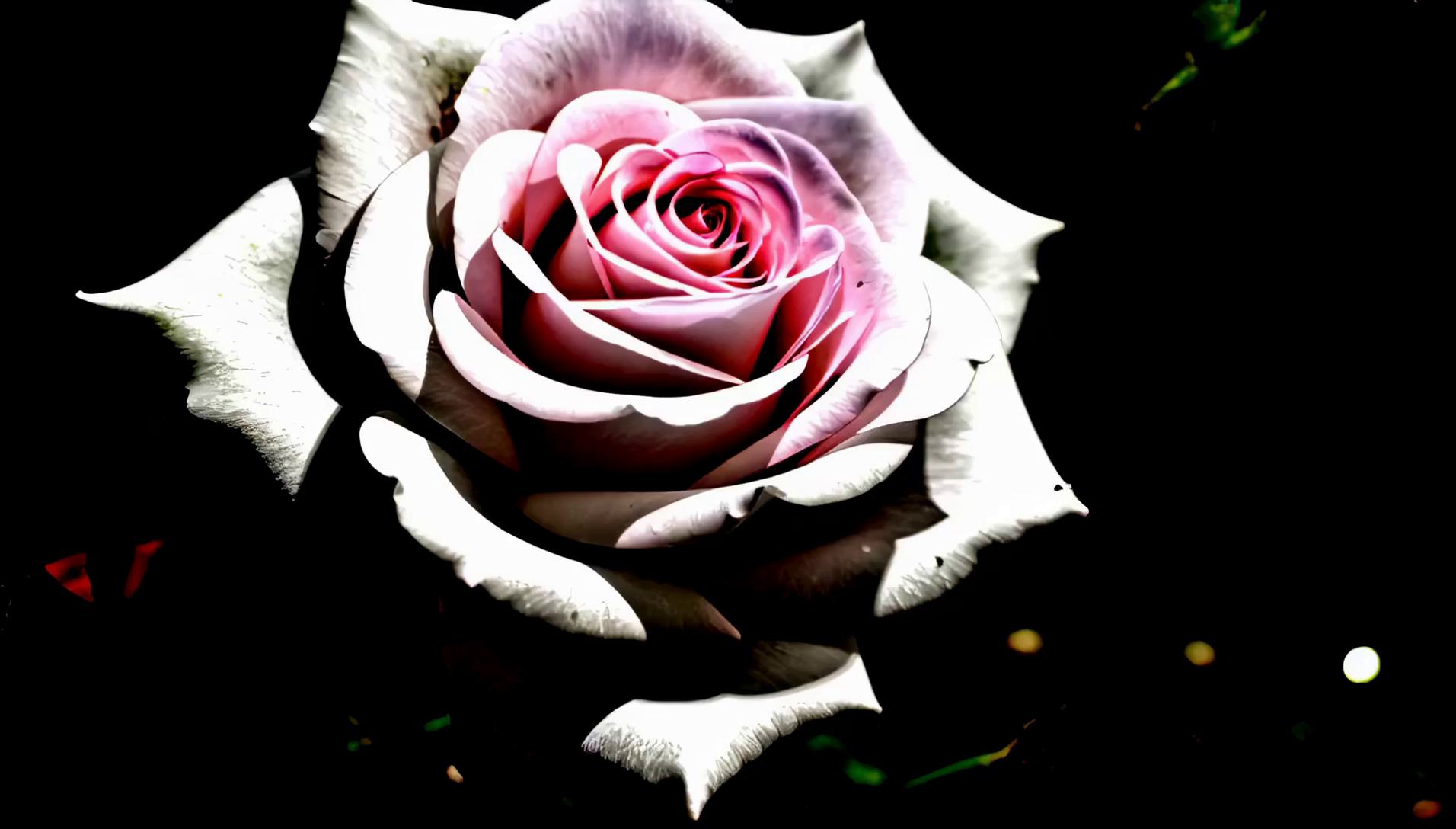}}
    \caption{\name{} can generate videos of all lengths up to 16 seconds, including images.}
    \label{fig:len}
\end{figure}

Since \name{} is trained on videos of various lengths, it can generate 1080p videos of all lengths up to 16 seconds, including images as videos of a single frame. We present examples in Figure~\ref{fig:len}.

\clearpage
\subsection{3D Consistency}

\begin{figure}[H]
    \centering
    \subfloat[Prompt: This portrait depicts an orange cat with blue eyes, slowly rotating, inspired by Vermeer's 'Girl with a Pearl Earring'. The cat is adorned with pearl earrings and has brown fur styled like a Dutch cap against a black background, illuminated by studio lighting.]{\includegraphics[width=\linewidth]{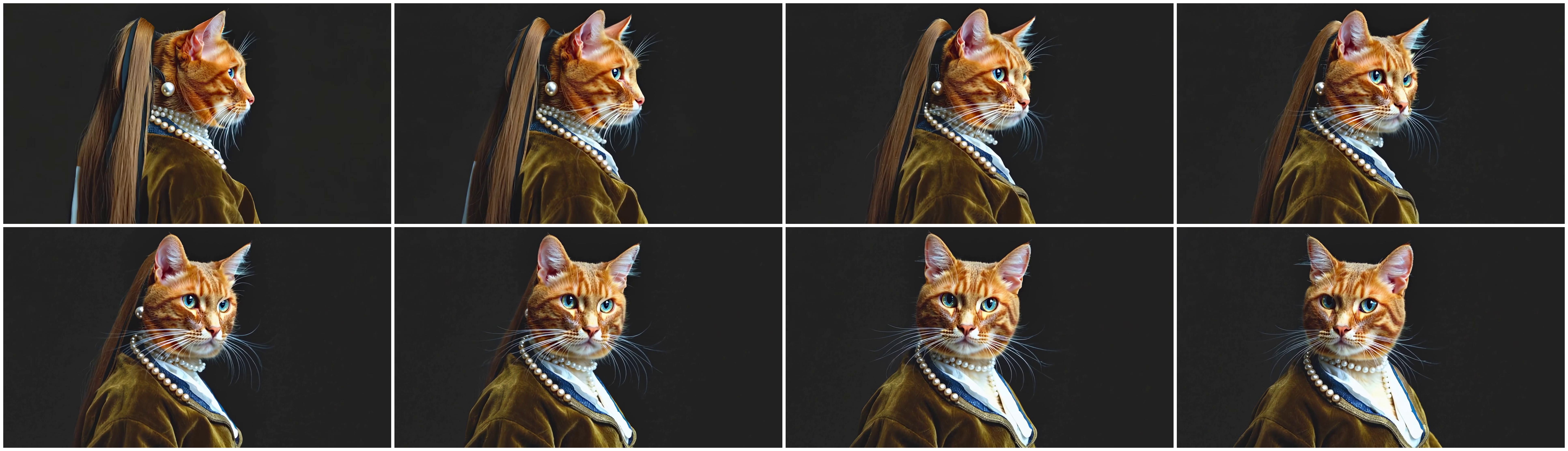}} \\
    \subfloat[Prompt: In a studio, there is a painting depicting a ship sailing through the rough sea.]{\includegraphics[width=\linewidth]{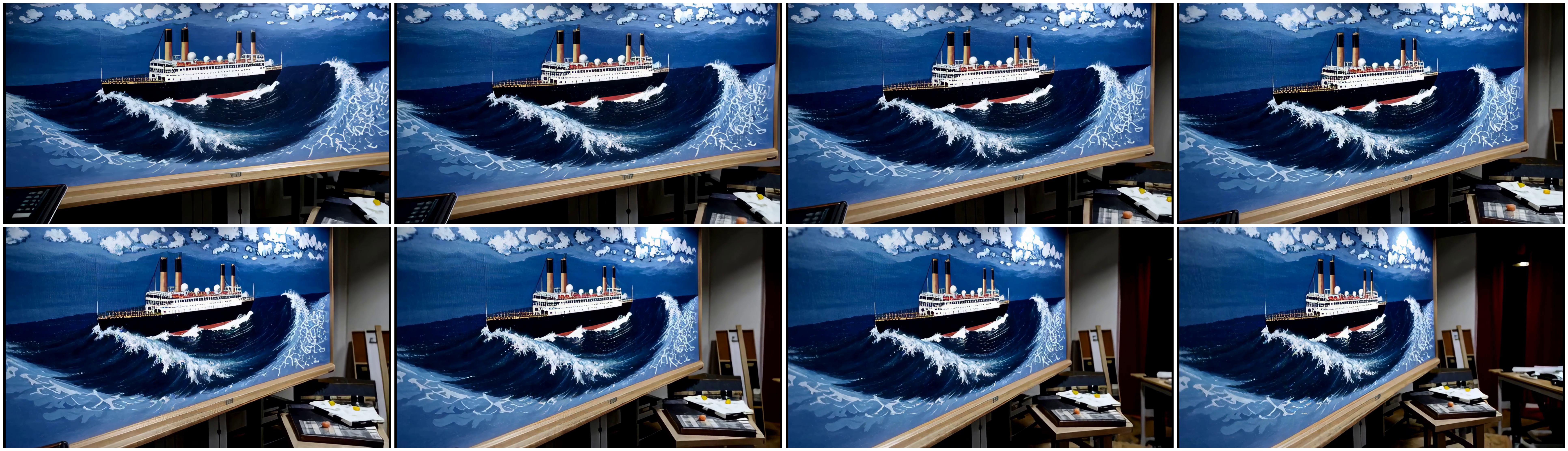}} \\
    \subfloat[Prompt: A red car is stuck in the snow, with the entire vehicle emitting green light and red signal lights flashing on the back. The camera slowly pans around the car.]{\includegraphics[width=\linewidth]{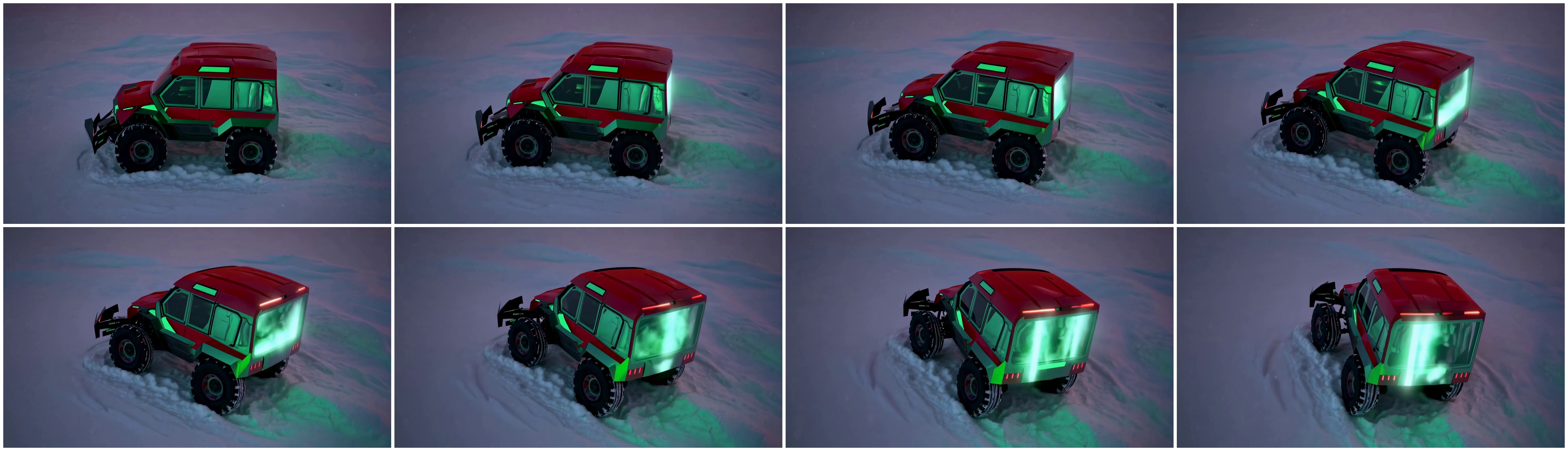}}
    \caption{3D consistency of \name{}.}
    \label{fig:3d}
\end{figure}

The video generated by \name{} exhibits strong 3D consistency. As the camera rotates, the video presents projections of the same object from different angles. For instance, as shown in Figure~\ref{fig:3d}, the hair of the generated cat naturally occludes as the camera rotates.

\clearpage
\subsection{Generating Cuts}
\begin{figure}[H]
    \centering
    \subfloat[Prompt: A sculptor is intently working on a clay bust, meticulously refining its facial features with precise hand movements.]
    {\includegraphics[width=\linewidth]{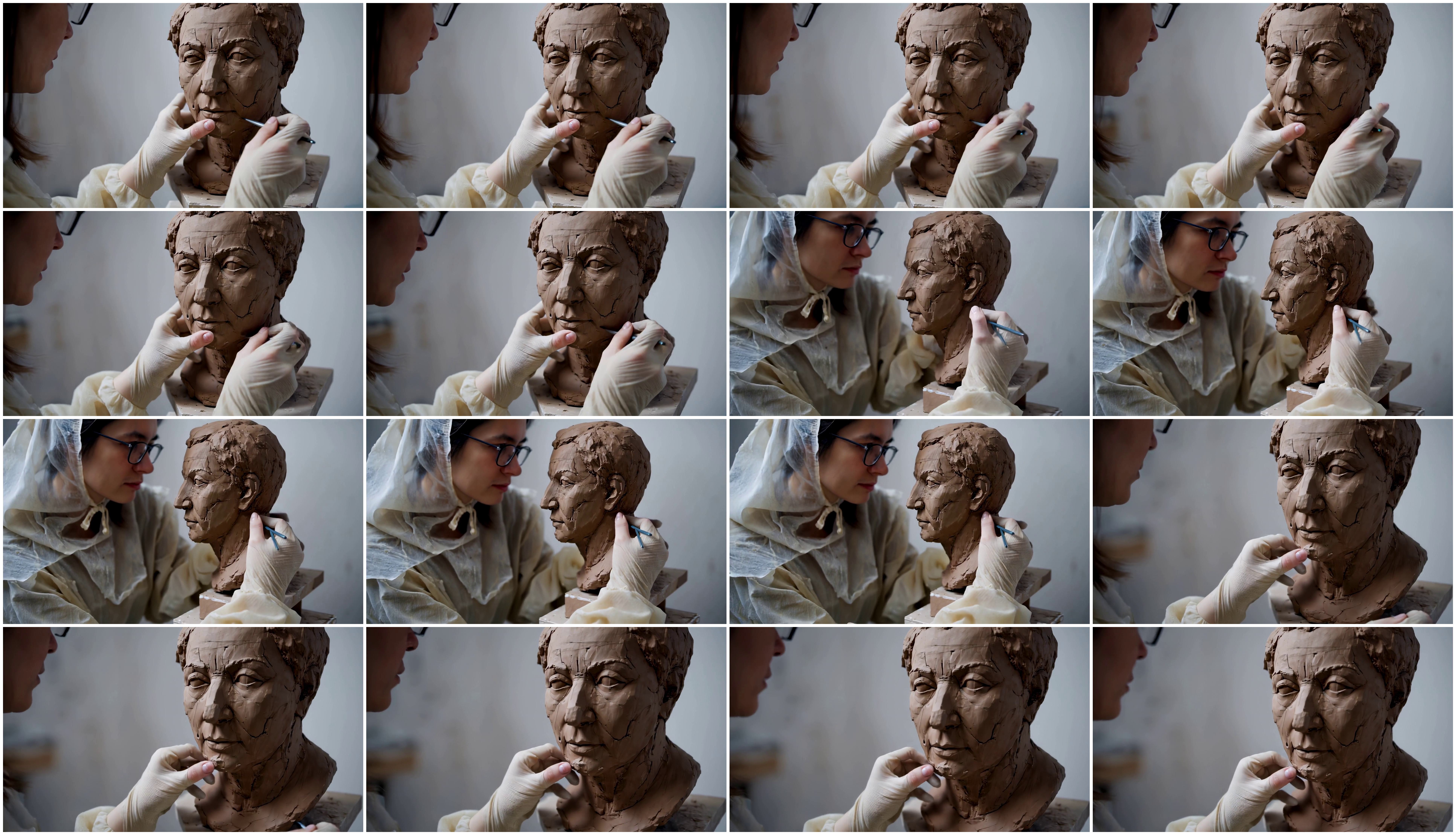}} \\
    \subfloat[Prompt: Churning ocean waves at night with a lighthouse on the coast create an intense and somewhat foreboding atmosphere.   The scene is set under an overcast sky, with the ocean's dark waters illuminated by natural light, highlighting the white foam of the waves.]{\includegraphics[width=\linewidth]{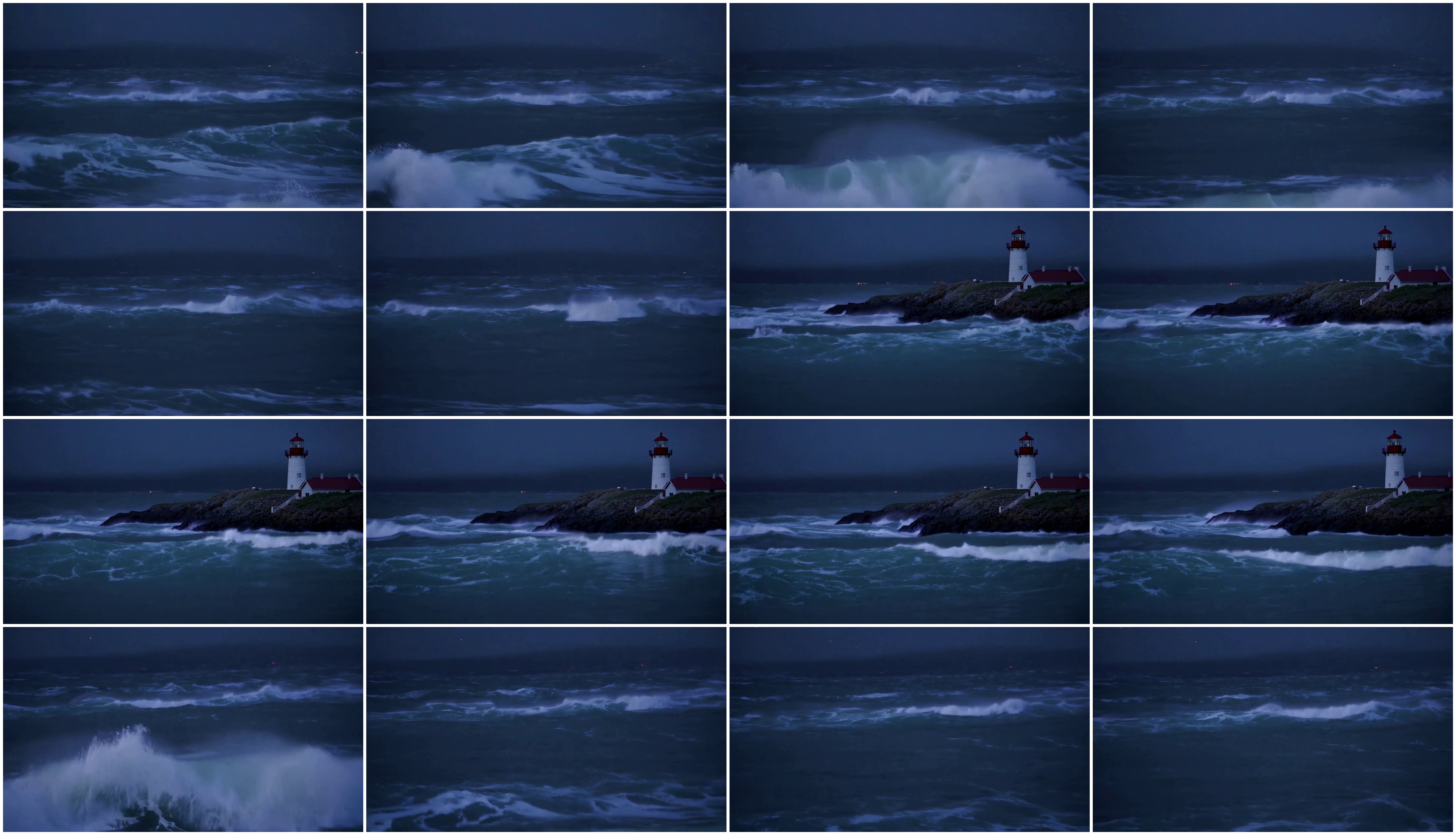}}
    \caption{\name{} is capable of generating videos with cuts.}
    \label{fig:cut}
\end{figure}

\name{} is capable of generating videos incorporating cuts. As shown in Figure~\ref{fig:cut}, these videos present different perspectives of the same scene by switching camera angles, while maintaining consistency of subjects in the scene.

\clearpage
\subsection{Generating Transitions}

\begin{figure}[H]
    \centering
    \subfloat[Prompt: An elderly man with glasses, dressed in formal attire, is deeply engrossed in examining a large, ornate pocket watch. As the video progresses, there is a cinematic transition to a fantastical mechanical cityscape, viewed through the openwork of the watch. This shift evokes a sense of wonder and transports the viewer into a steampunk-inspired world where buildings and structures are made of metal and gears.]
    {\includegraphics[width=\linewidth]{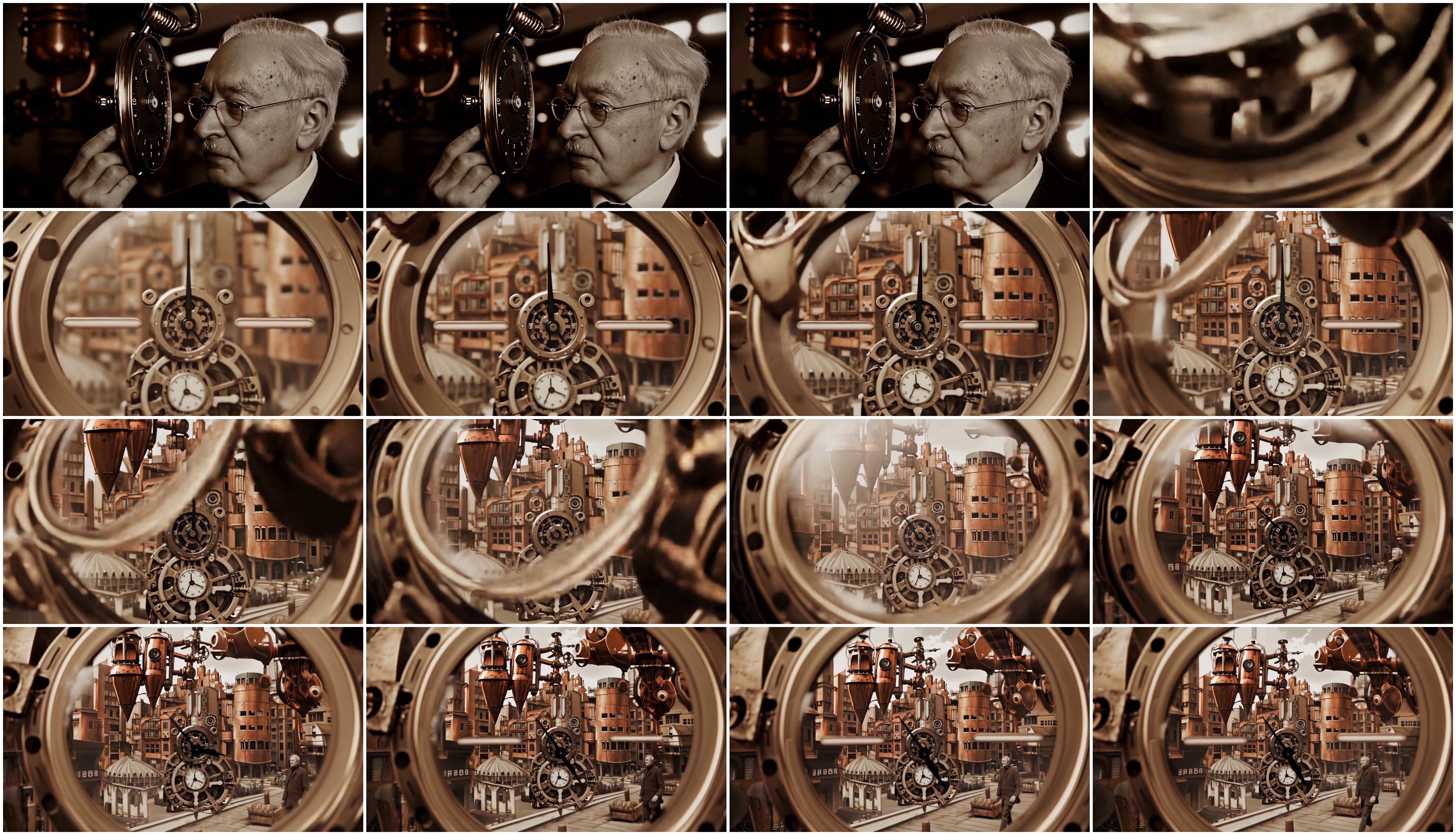}} \\
    \subfloat[Prompt: A person holding a dessert with a fluffy layer of whipped cream elegantly drizzled with smooth chocolate sauce. As a dollop of cream falls, a mini polar bear appears, with floating icebergs nearby, set against a serene blue backdrop.]{\includegraphics[width=\linewidth]{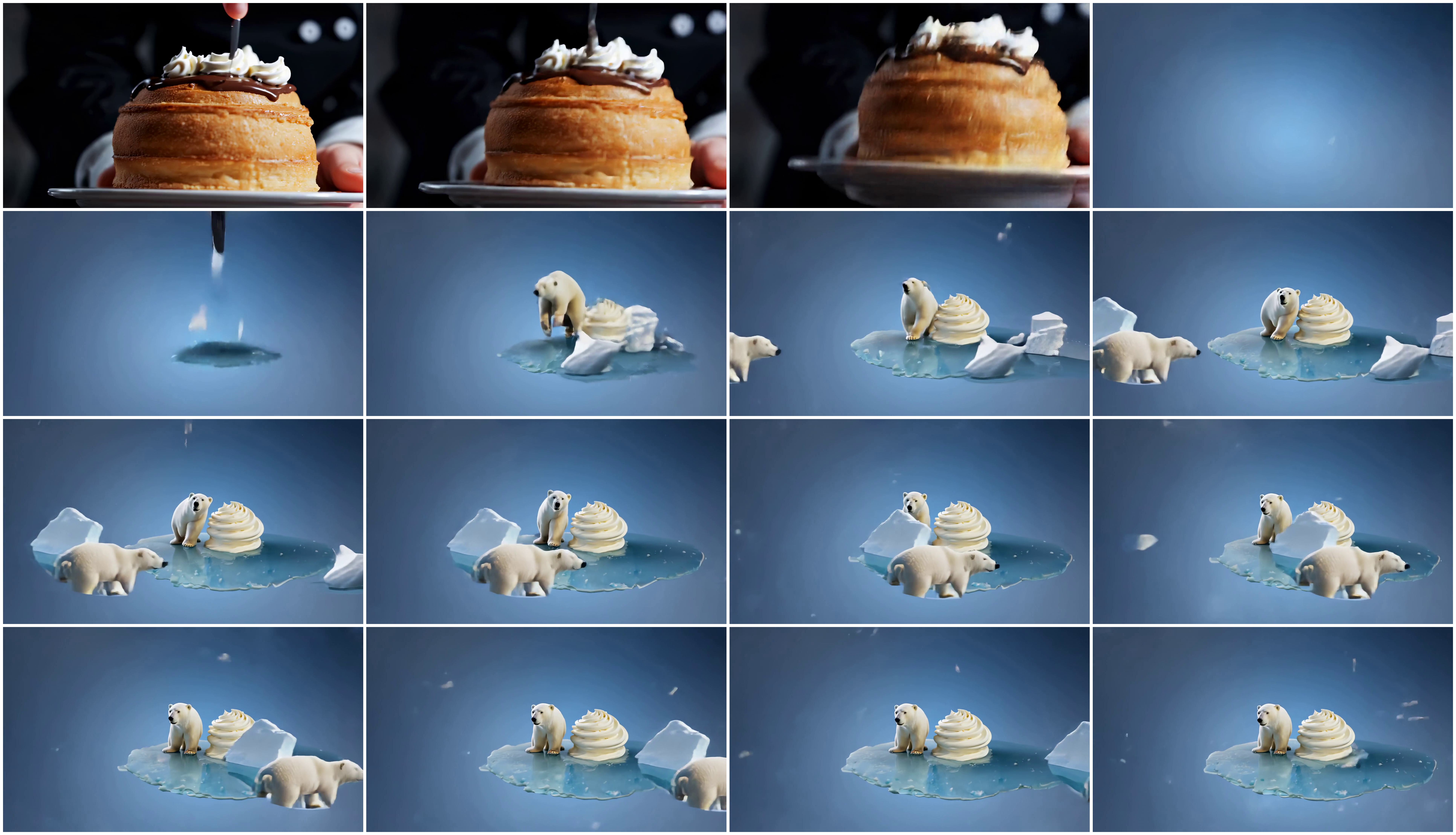}}
    \caption{\name{} is capable of generating videos with transitions.}
    \label{fig:st}
\end{figure}

\name{} is capable of producing videos with transitions in a single generation. As shown in Figure~\ref{fig:st}, these transitions can connect two different scenes in an engaging manner.

\clearpage
\subsection{Camera Movements}
\begin{figure}[H]
    \centering
    \subfloat[Zoom. Prompt: A large sailing ship sails slowly through the fog.]{\includegraphics[width=\linewidth]{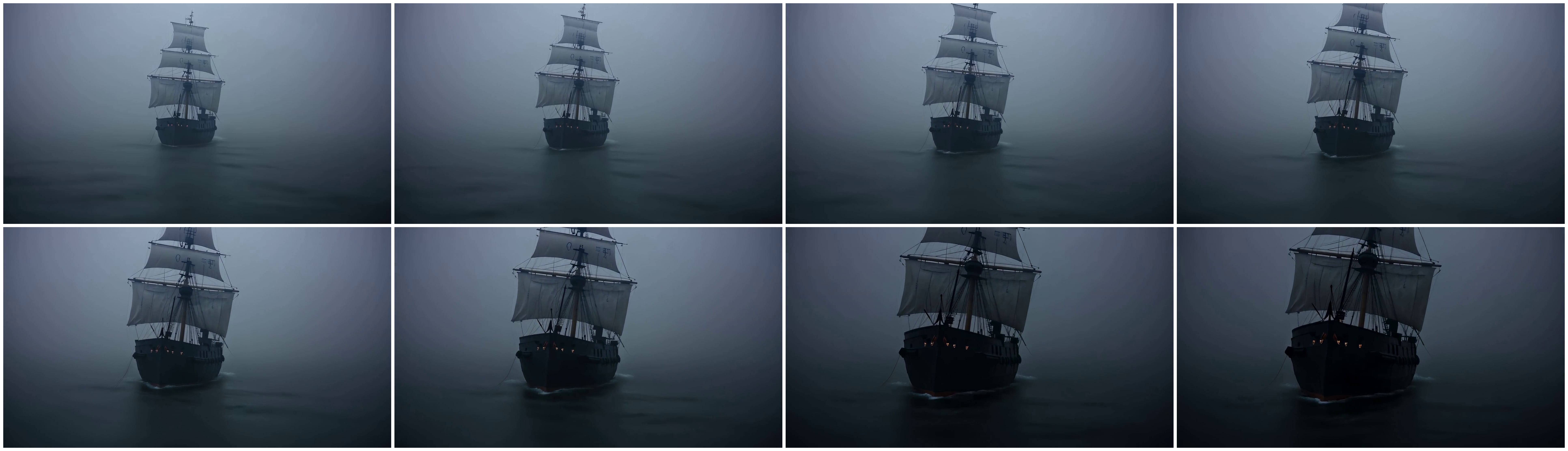}} \\
    \subfloat[Pan. Prompt: An elderly man with a white beard is seated in a room filled with wooden bookshelves, brimming with old books.  He is dressed in a dark suit and tie, and he is engrossed in reading a large book.  The room is bathed in the warm glow of sunlight streaming through a window, creating a serene and contemplative atmosphere.]{\includegraphics[width=\linewidth]{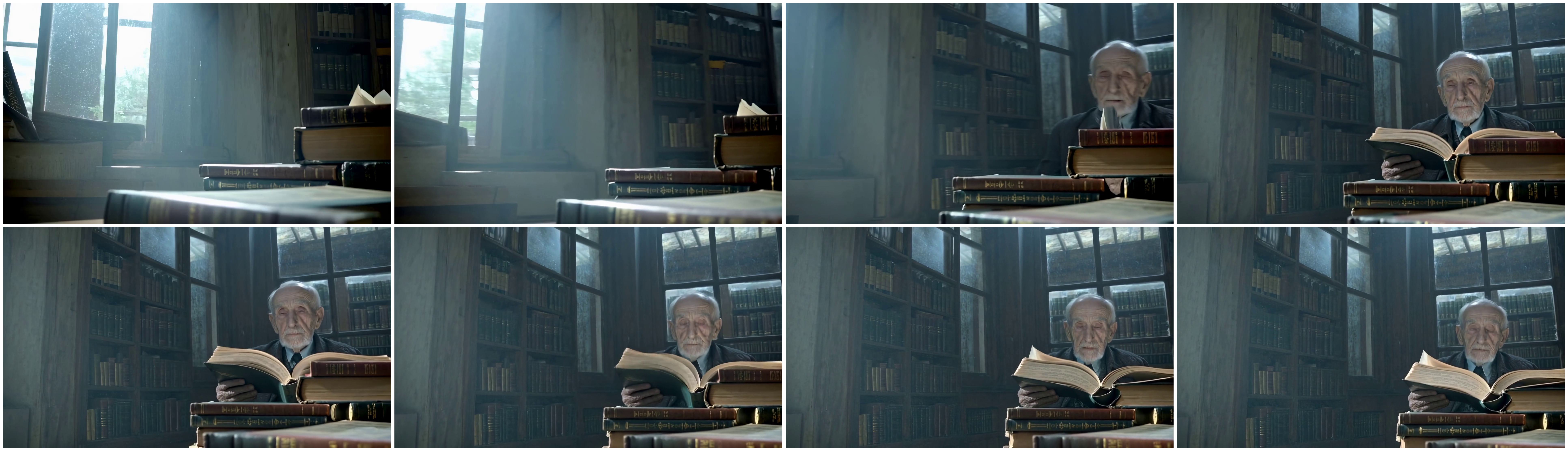}} \\
    \subfloat[Dolly. Prompt: An animated hedgehog with distinctive spiky hair and large eyes is seen exploring a lush, grassy environment.]{\includegraphics[width=\linewidth]{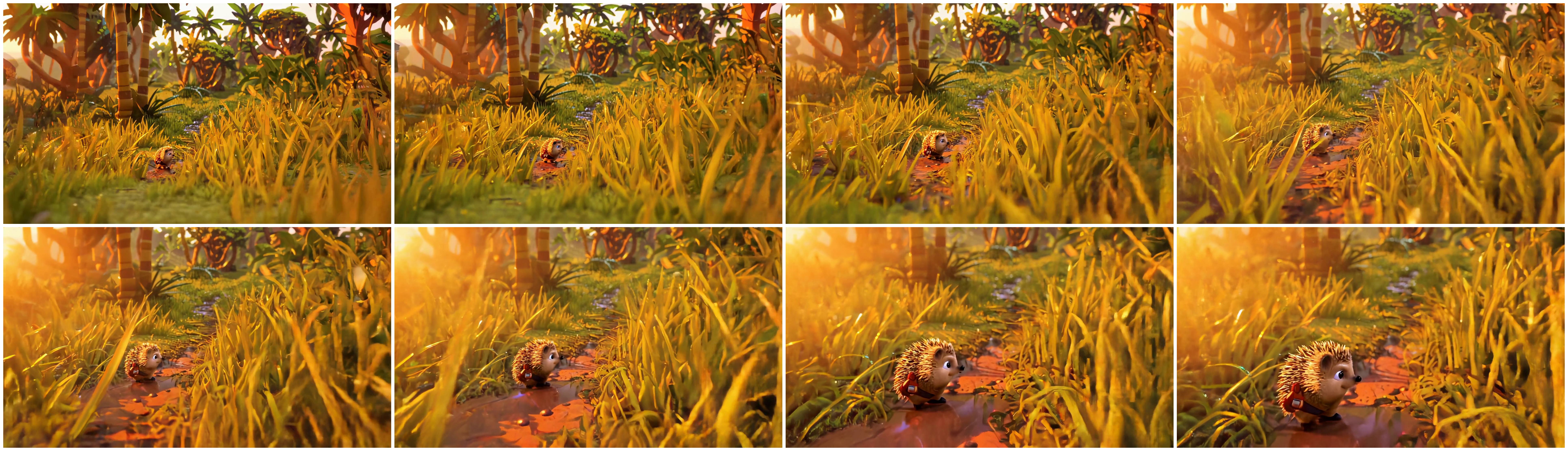}} \\
    \caption{Camera movements generated by \name{}.}
    \label{fig:cm}
\end{figure}

Camera movements involve the physical adjustments or movements of a camera during filming, enhancing visual narrative and conveying various perspectives and emotions within scenes. \name{} learned these techniques from the data, enhancing the visual experience of viewers.
For instance, as shown in Figure~\ref{fig:cm}, \name{} is capable of generating videos with camera movements including zoom, pan and dolly.

\clearpage
\subsection{Lighting Effects}

\begin{figure}[H]
    \centering
    \subfloat[Prompt: A man wearing a hat and a dark suit walks from the corridor towards the room.  The lighting casts a bluish tint over the scene, creating a suspenseful atmosphere.]
    {\includegraphics[width=\linewidth]{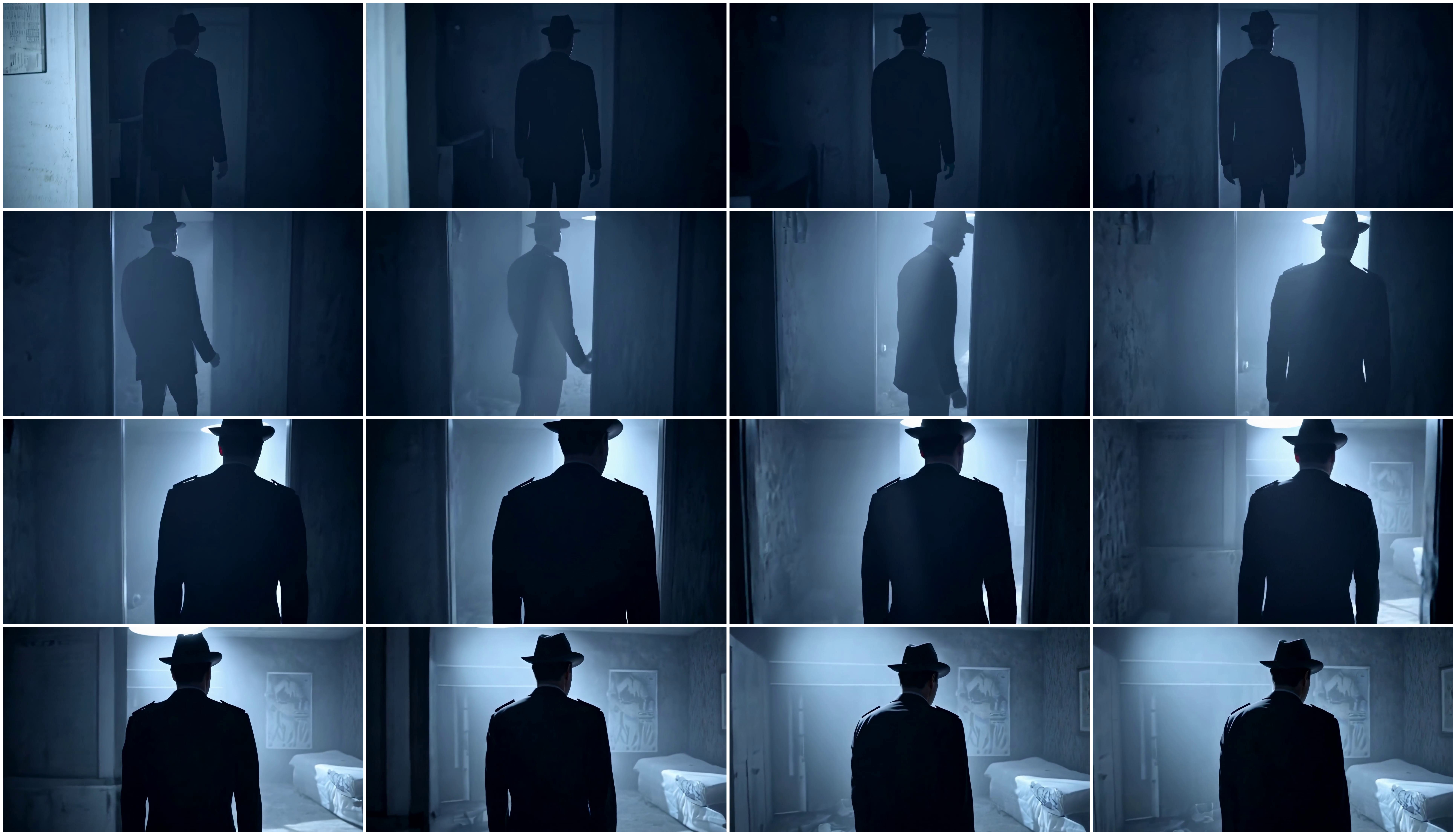}} \\
    \subfloat[Prompt: A rustic wooden cabin nestles by the shore of a clear, sunlit lake, surrounded by verdant trees and mountains. The water is calm, reflecting the sky above, with a few clouds scattered across it. Sailboats and kayaks are moored on the lake, inviting leisure and tranquility.]{\includegraphics[width=\linewidth]{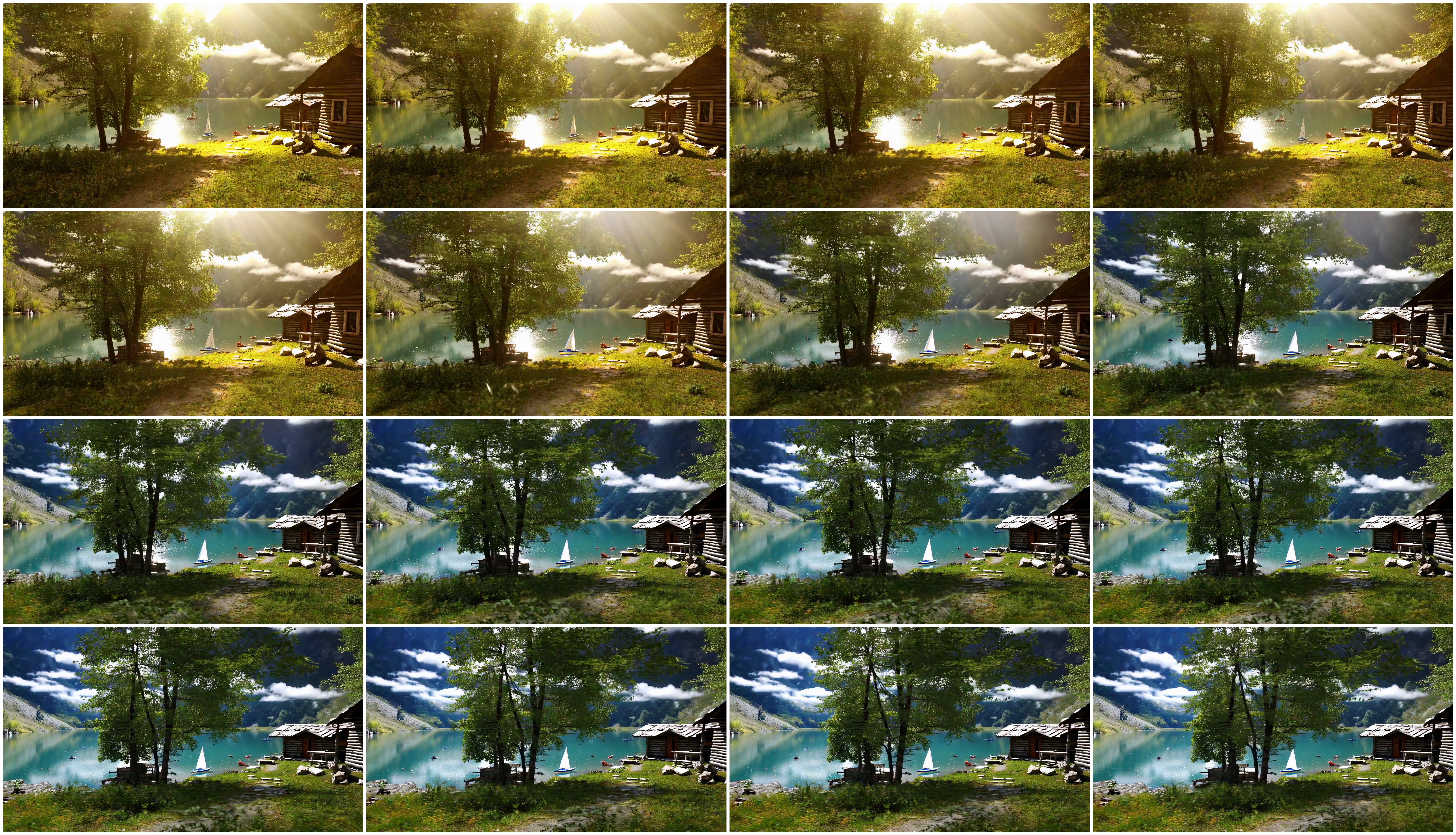}}
    \caption{Lighting effects generated by \name{}.}
    \label{fig:light}
\end{figure}

\name{} is capable of generating videos with impressive lighting effects, which help enhance the overall atmosphere. For example, as shown in Figure~\ref{fig:light}, the generated videos can evoke atmospheres of mystery and tranquility. Therefore, besides the entities within the video content, \name{} has the preliminary ability to convey some abstract feelings.

\clearpage
\subsection{Emotional Portrayal}

\begin{figure}[H]
    \centering
    \subfloat[Prompt: A man and a woman are sharing a close and affectionate interaction in an indoor setting that suggests a romantic ambiance.]
    {\includegraphics[width=\linewidth]{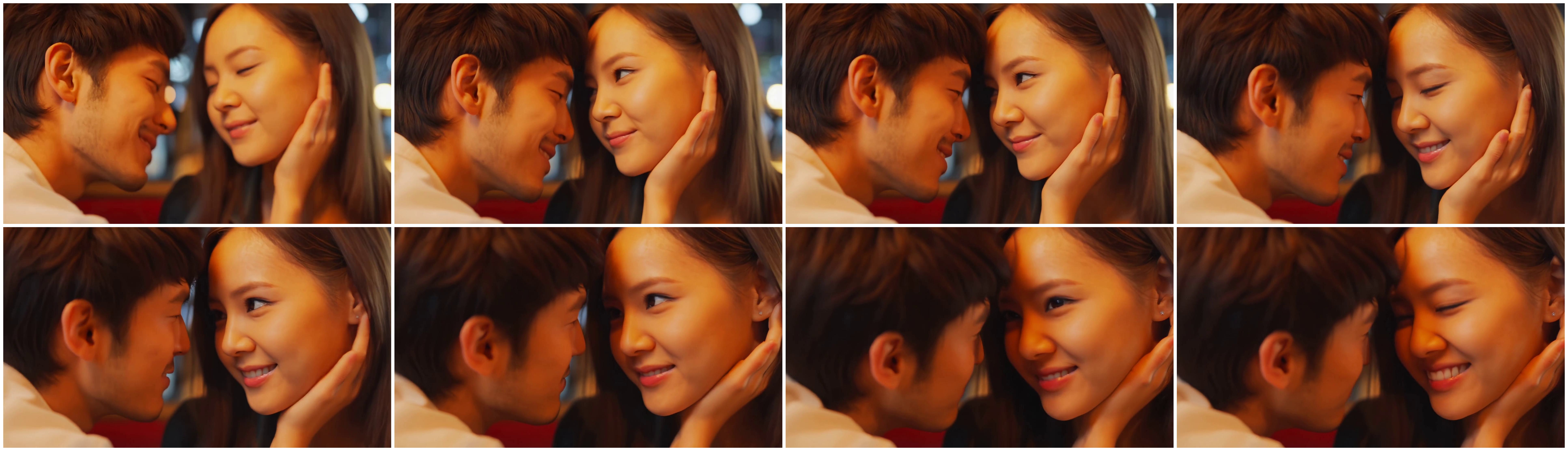}} \\
    \subfloat[Prompt: An elderly woman with white hair and a lined face is seated inside an older model car, looking out through the side window with a contemplative or mildly sad expression.]{\includegraphics[width=\linewidth]{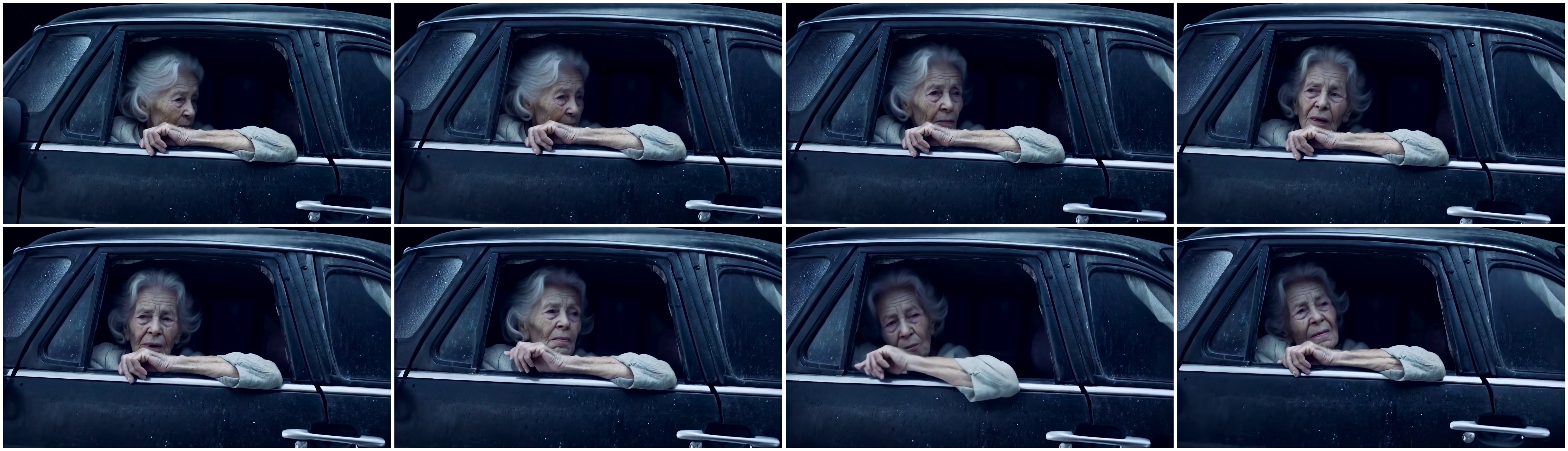}} \\
    \subfloat[Prompt: A couple about to get divorced sat awkwardly in the waiting room.]{\includegraphics[width=\linewidth]{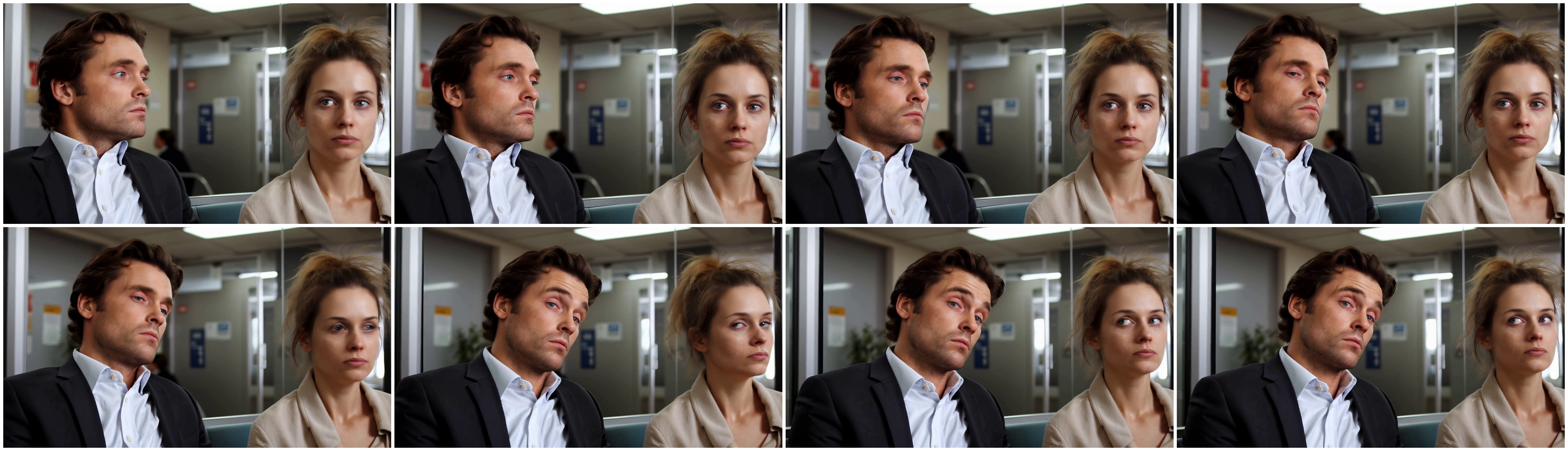}} \\
    \subfloat[Prompt: Audience members in a theater are captured in a series of medium shots, with a young man and woman in formal attire centrally positioned and illuminated by a spotlight effect.]{\includegraphics[width=\linewidth]{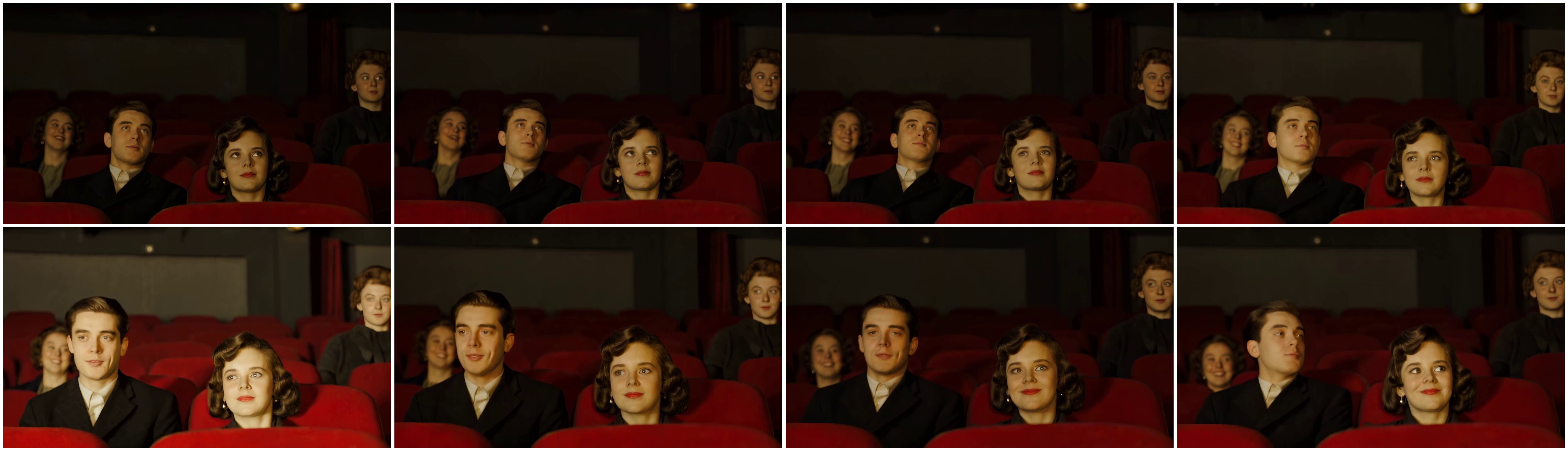}}
    \caption{Emotional portrayal of \name{}.}
    \label{fig:emo}
\end{figure}

\name{} is able to depict characters' emotions effectively. For example, as shown in Figure~\ref{fig:emo}, \name{} can express emotions such as happiness, loneliness, embarrassment, and joy.

\clearpage
\subsection{Imaginative Ability}

\begin{figure}[H]
    \centering
    \subfloat[Prompt: A painting of a boat on water comes to life, with waves crashing and the boat becoming submerged.]
    {\includegraphics[width=\linewidth]{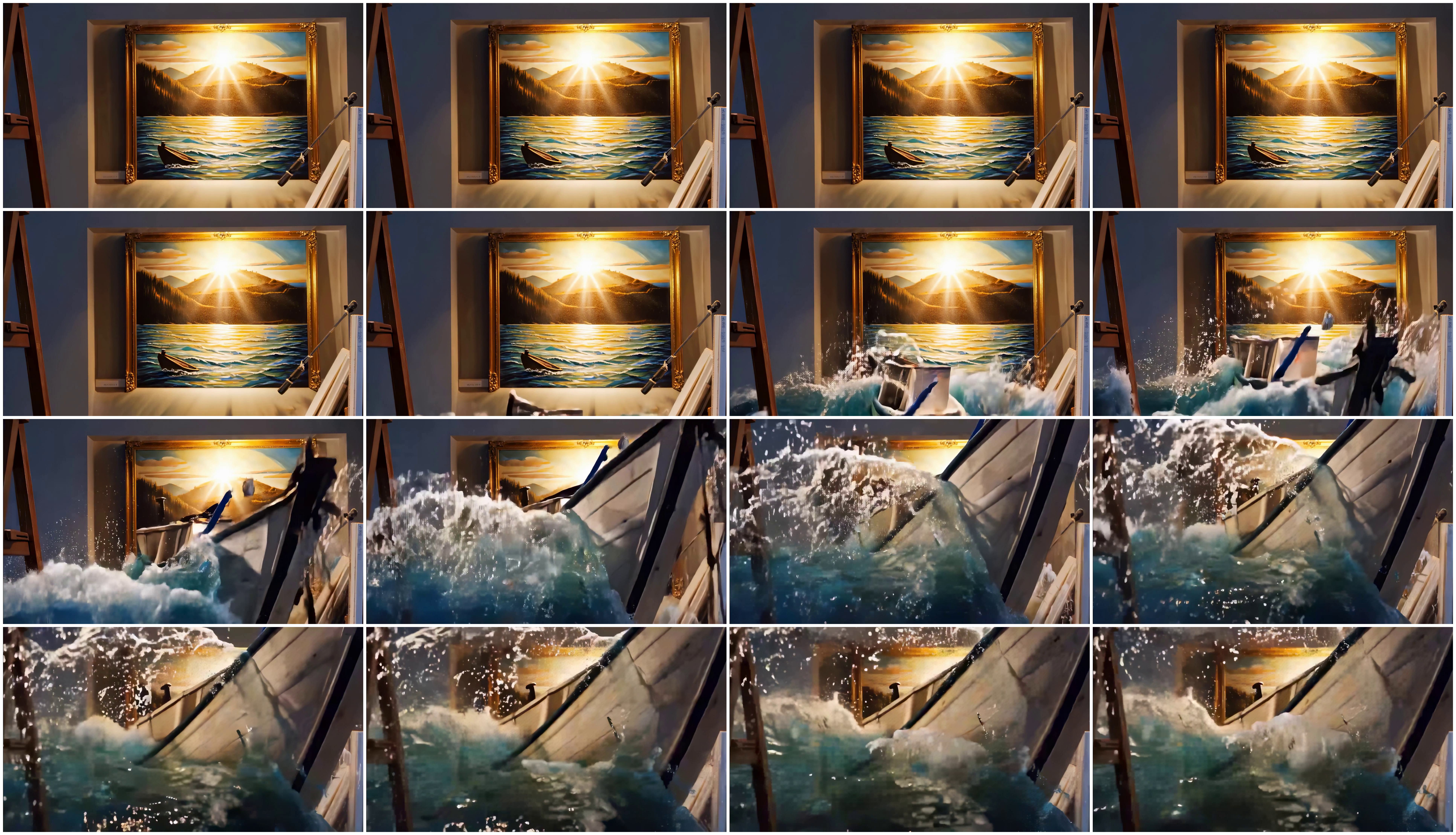}} \\
    \subfloat[Prompt: An animated rabbit in a playful pink snowboarding outfit is carving its way down a snowy mountain slope under a clear blue sky.]{\includegraphics[width=\linewidth]{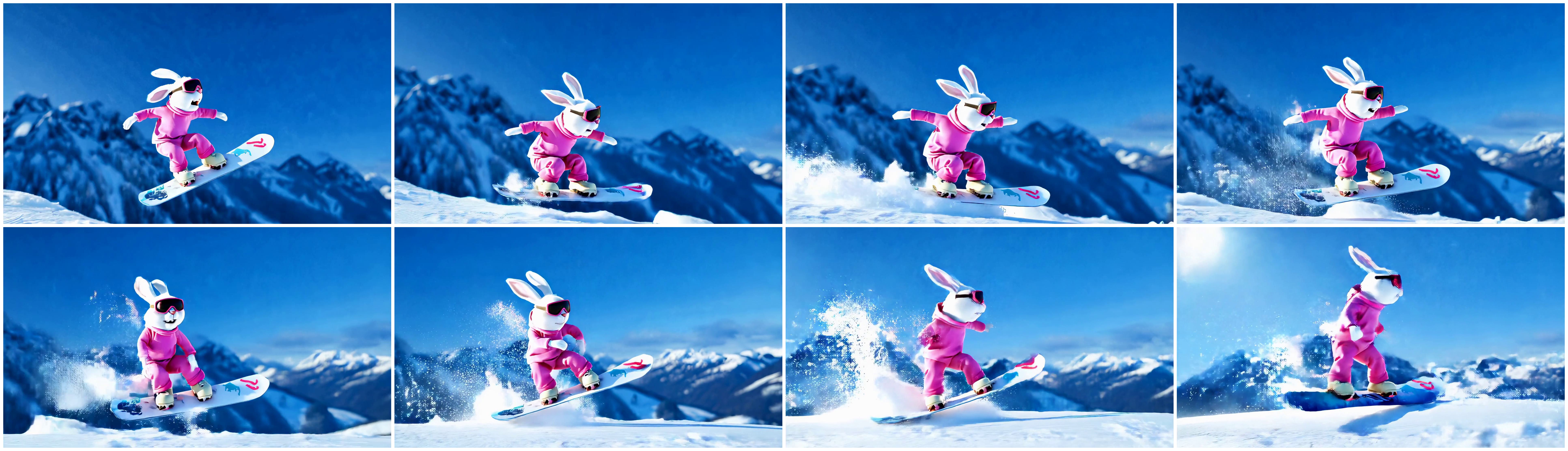}} \\
    \subfloat[Prompt: A model train with a blue engine is seen traveling through a meticulously crafted miniature landscape. The train is pulling several red and cream-colored passenger cars along a track that winds through a rural or suburban setting with small-scale houses, verdant trees, and miniature waterfalls.]{\includegraphics[width=\linewidth]{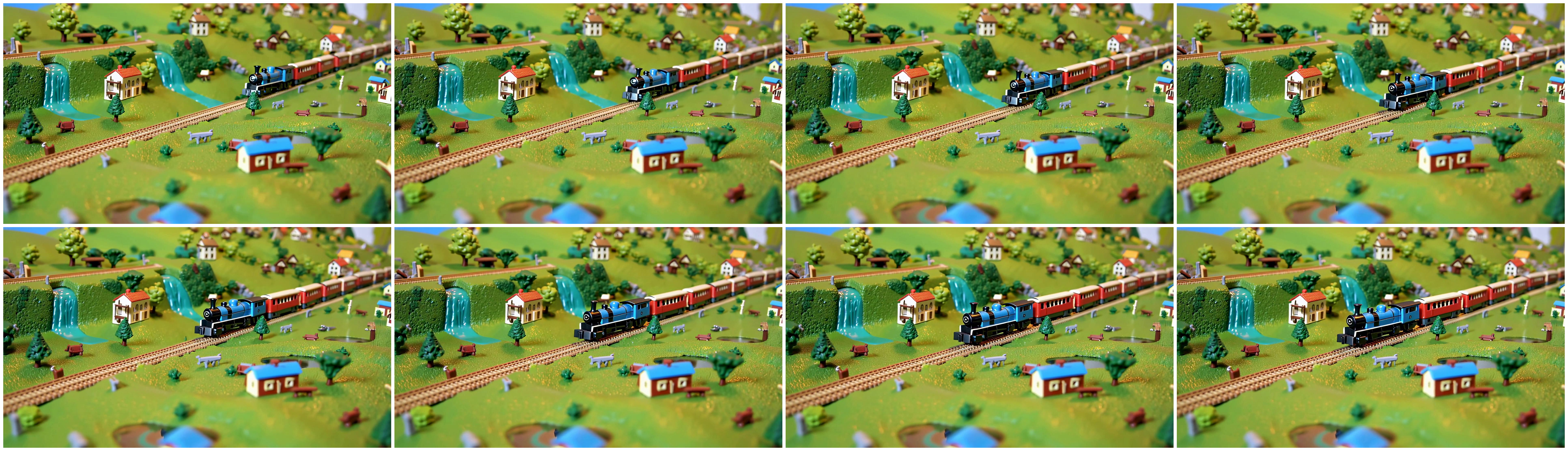}}
    \caption{Imaginative ability of \name{}.}
    \label{fig:ia}
\end{figure}

In addition to generating real-world scenes, \name{} also possesses a rich imagination. As shown in Figure~\ref{fig:ia}, \name{} is able to generate scenes that do not exist in the real world.

\clearpage
\subsection{Comparison with Sora}

\begin{figure}[H]
    \centering
    \subfloat[Sora]
    {\includegraphics[width=\linewidth]{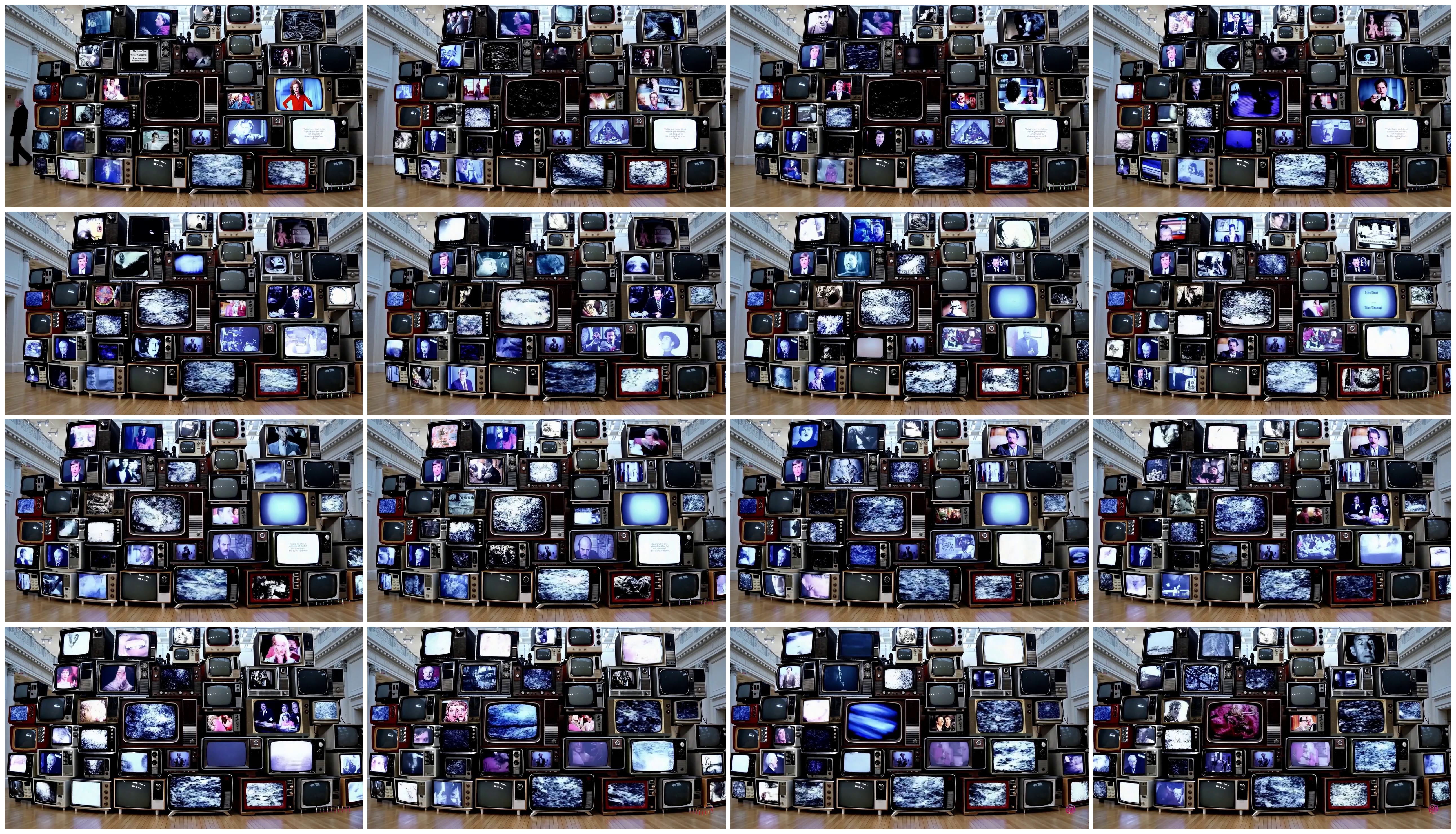}} \\
    \subfloat[Vidu]{\includegraphics[width=\linewidth]{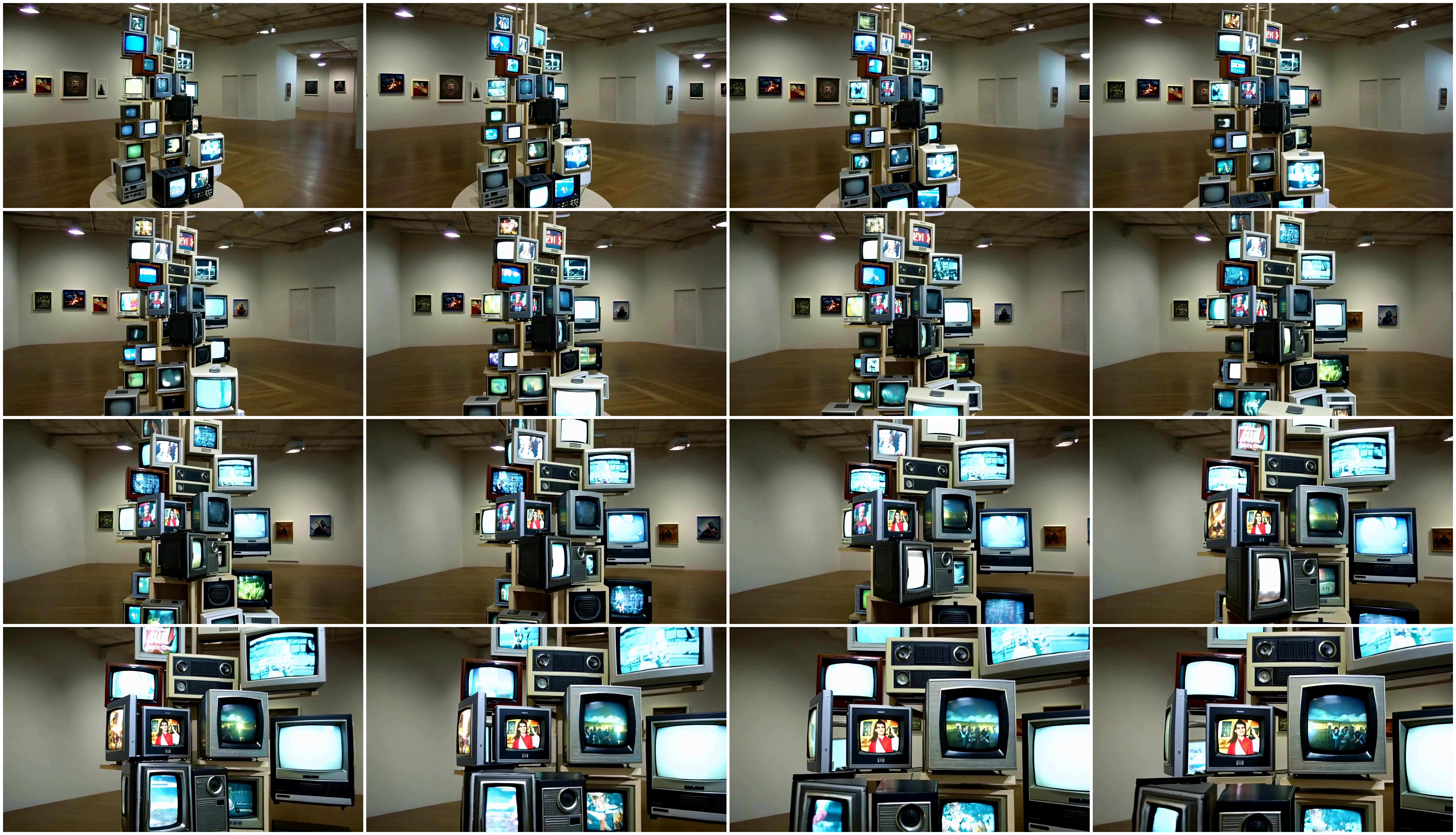}}
    \caption{Prompt: The camera rotates around a large stack of vintage televisions all showing different programs — 1950s sci-fi movies, horror movies, news, static, a 1970s sitcom, etc, set inside a large New York museum gallery.}
    \label{fig:tv}
\end{figure}

\begin{figure}[H]
    \centering
    \subfloat[Sora]
    {\includegraphics[width=\linewidth]{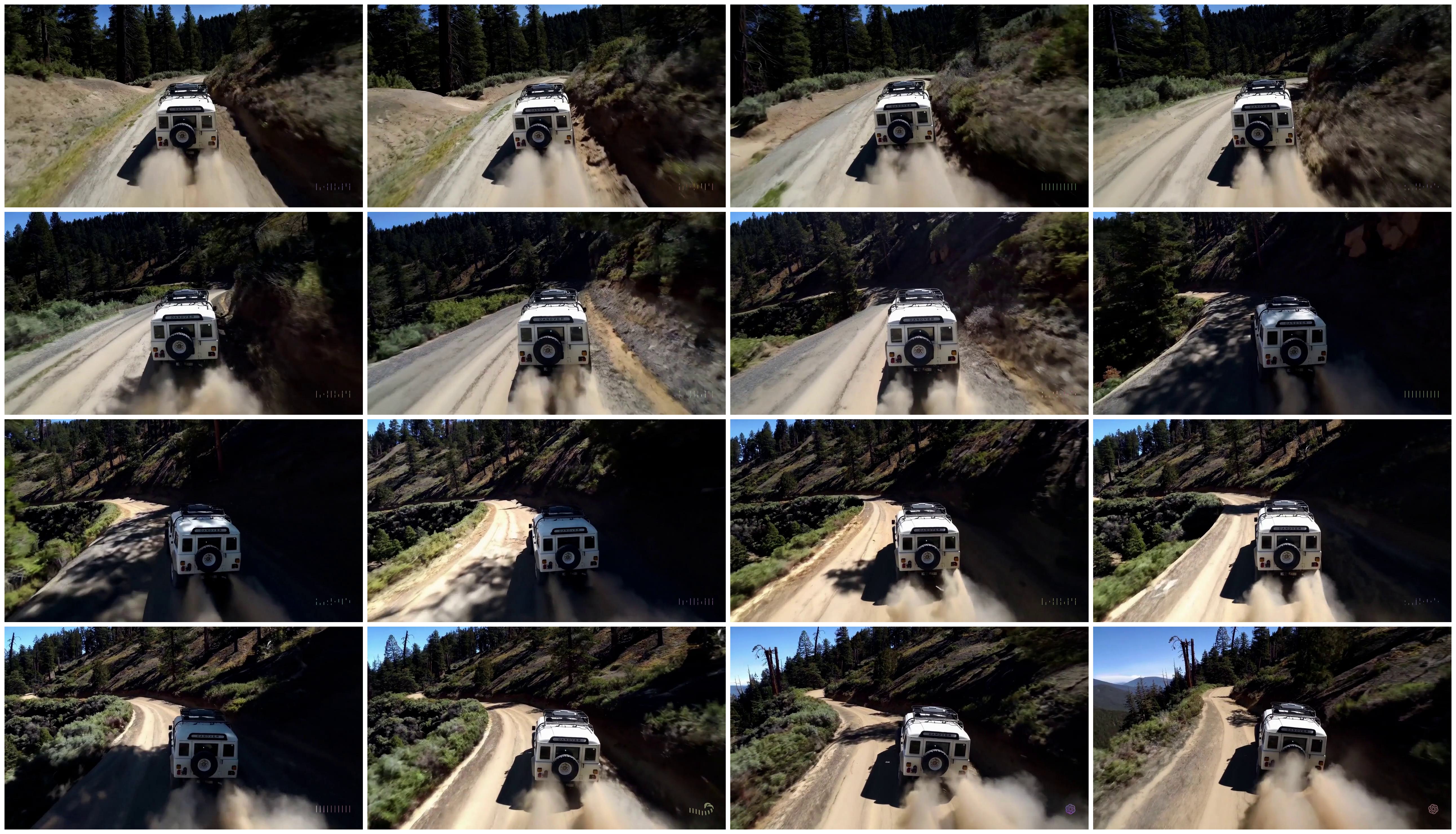}} \\
    \subfloat[Vidu]{\includegraphics[width=\linewidth]{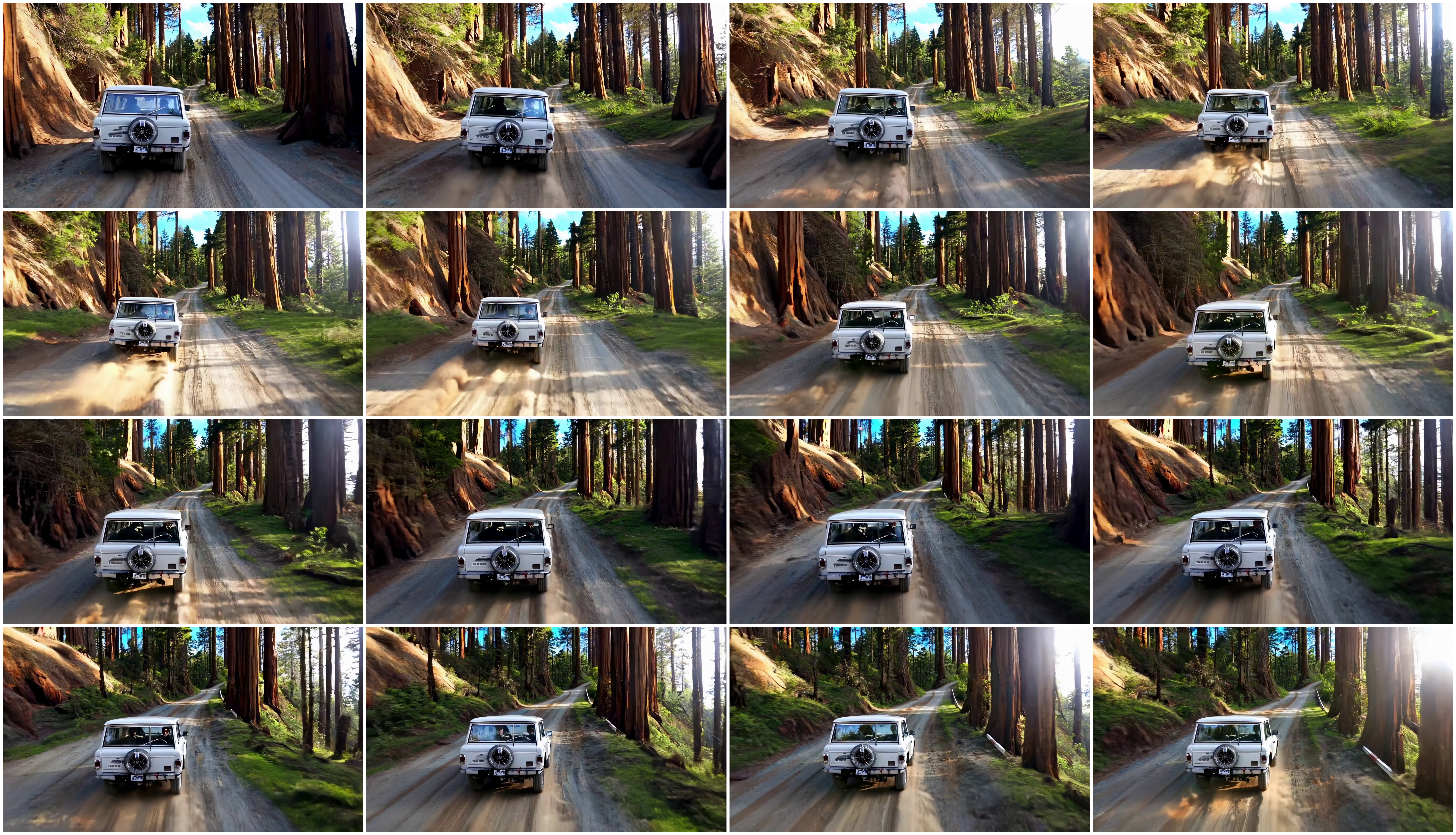}}
    \caption{Prompt: The camera follows behind a white vintage SUV with a black roof rack as it speeds up a steep dirt road surrounded by pine trees on a steep mountain slope, dust kicks up from it’s tires, the sunlight shines on the SUV as it speeds along the dirt road, casting a warm glow over the scene. The dirt road curves gently into the distance, with no other cars or vehicles in sight. The trees on either side of the road are redwoods, with patches of greenery scattered throughout. The car is seen from the rear following the curve with ease, making it seem as if it is on a rugged drive through the rugged terrain. The dirt road itself is surrounded by steep hills and mountains, with a clear blue sky above with wispy clouds.}
    \label{fig:car}
\end{figure}

Sora~\cite{videoworldsimulators2024} is currently the most powerful text-to-video generator, capable of producing high-definition videos with high consistency. However, as Sora is not publicly accessible, we compare them by inserting the example prompts released by Sora directly to Vidu. Figure~\ref{fig:tv} 
 and Figure~\ref{fig:car} illustrate the comparison between \name{} and Sora, indicating that to some extent, the generation performance of \name{} is comparable to Sora.

\clearpage

\clearpage
\section{Other Controllable Video Generation}

We also perform several initial experiments at 512 resolution on other controllable video generation, including canny-to-video generation~\cite{zhang2023adding}, video prediction, and subject-driven generation~\cite{ruiz2023dreambooth}. All of them demonstrate promising results.

\subsection{Canny-to-Video Generation}

\begin{figure}[H]
    \centering
    \subfloat[Input canny.]{\includegraphics[width=\linewidth]{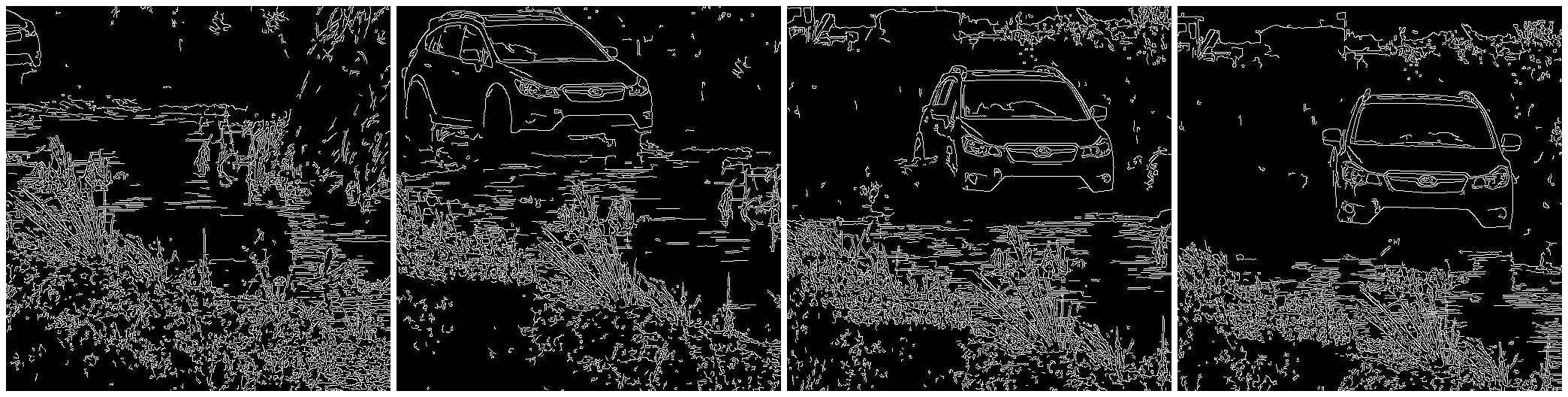}} \\
    \subfloat[Prompt: During the day, a white car drove towards me and splashed water as it passed by a pond, realistic visual style.]{\includegraphics[width=\linewidth]{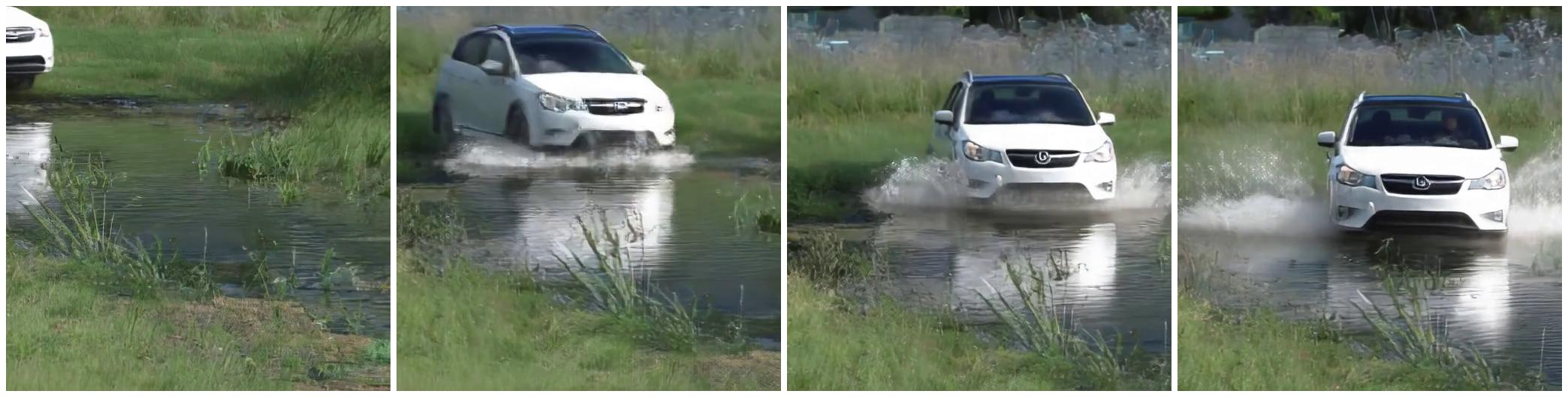}} \\
    \subfloat[Prompt: During the day, a red car drove towards me and splashed water as it passed by a pond, realistic visual style.]{\includegraphics[width=\linewidth]{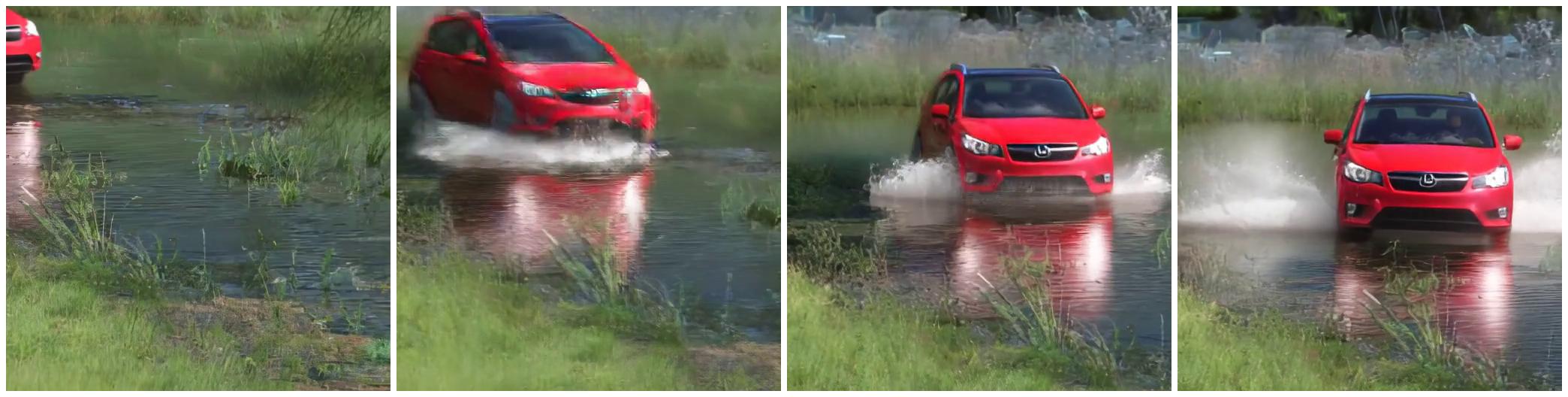}} \\
    \subfloat[Prompt: During the day, a white car drove towards me and splashed water as it passed by a pond, anime style.]{\includegraphics[width=\linewidth]{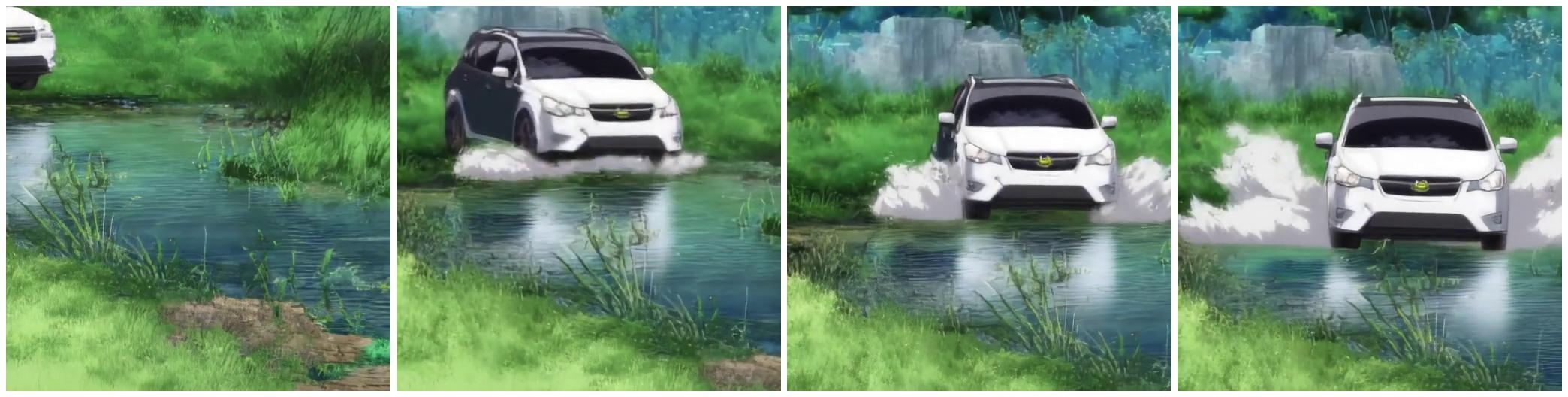}} \\
    \caption{Canny-to-video generation examples of \name{}.}
    \label{fig:c2v}
\end{figure}

\name{} can add additional control by using techniques similar to ControlNet~\cite{zhang2023adding}, as shown in Figure~\ref{fig:c2v}.

\clearpage
\subsection{Video Prediction}

\begin{figure}[H]
    \centering
    \subfloat[Prompt: A pink chrysanthemum flower with intricate petals is the focal point, resting on a wooden surface in an indoor setting.]{\includegraphics[width=\linewidth]{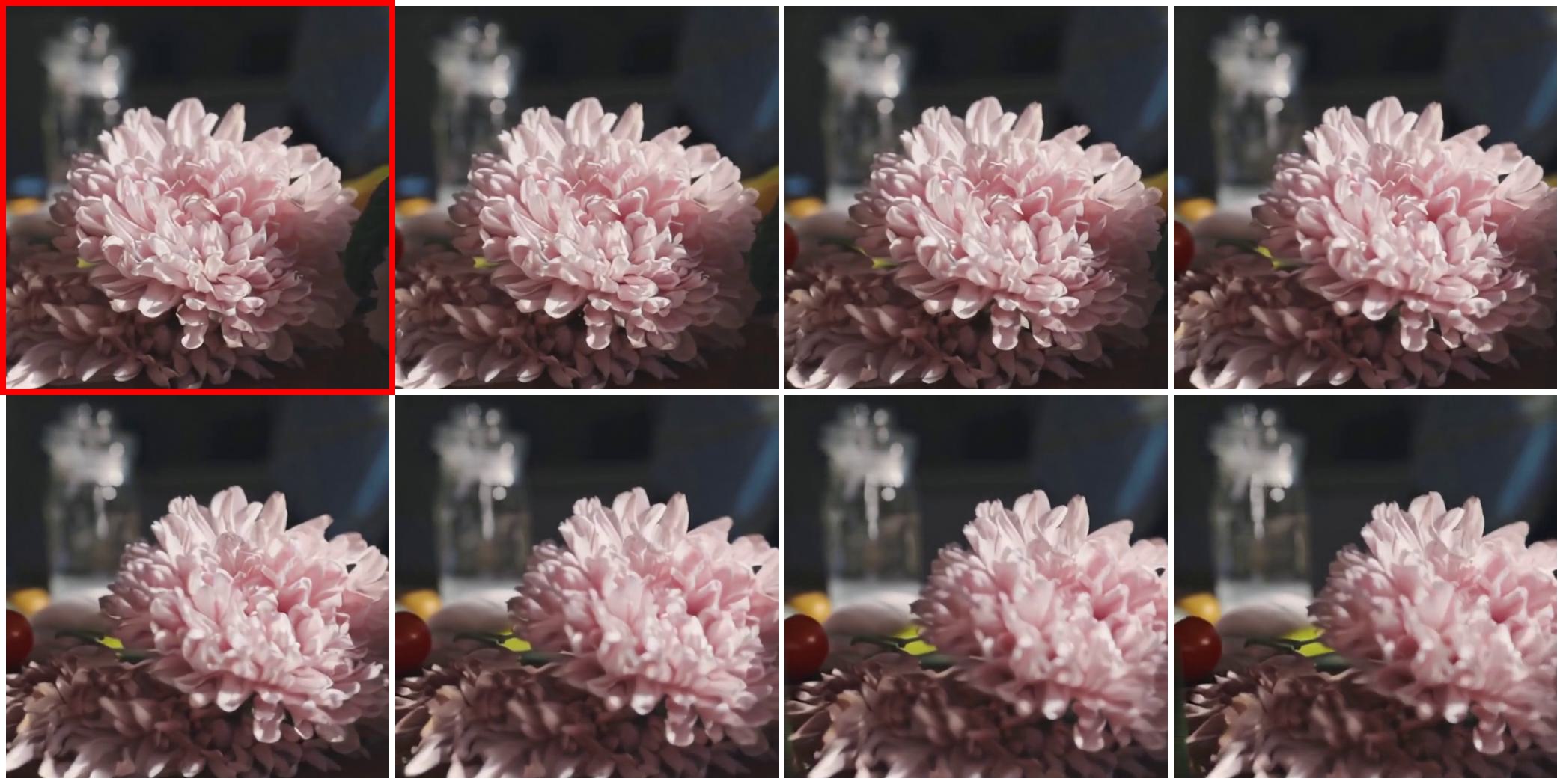}} \\
    \subfloat[Prompt: A serene mountainous landscape bathed in the warm glow of sunset or twilight, with snow-capped peaks rising above the green vegetation-covered slopes.  A calm body of water rests in the foreground, reflecting the sky above, which is dotted with clouds tinged with pink and orange hues.]{\includegraphics[width=\linewidth]{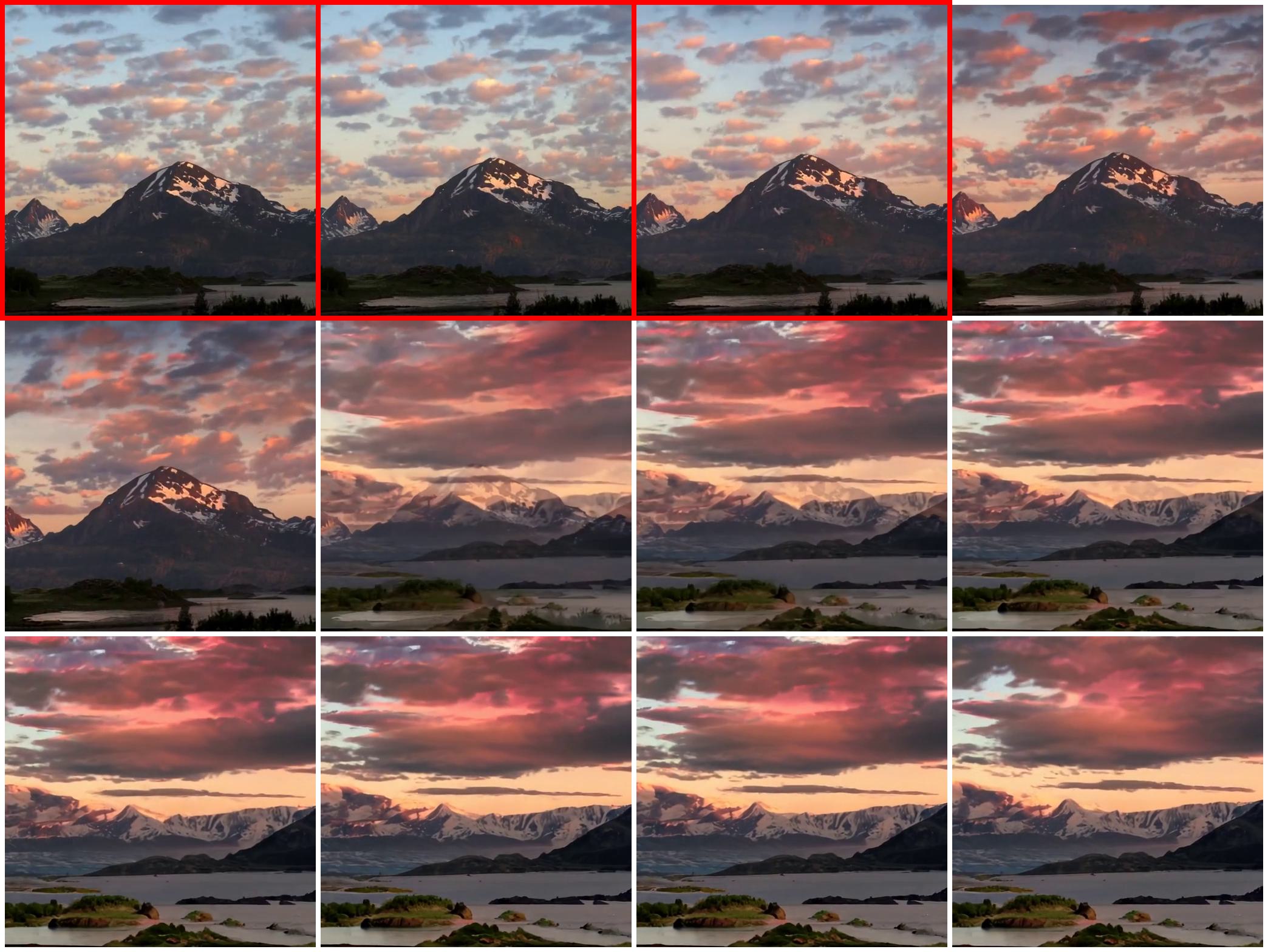}} \\
    \caption{Video prediction examples of \name{}.}
    \label{fig:vp}
\end{figure}

As shown in Figure~\ref{fig:vp}, \name{} can generate subsequent frames, given an input image, or several input frames (marked with red boxes).

\clearpage
\subsection{Subject-Driven Generation}

\begin{figure}[H]
    \centering
    \subfloat[Input images.]{\includegraphics[width=\linewidth]{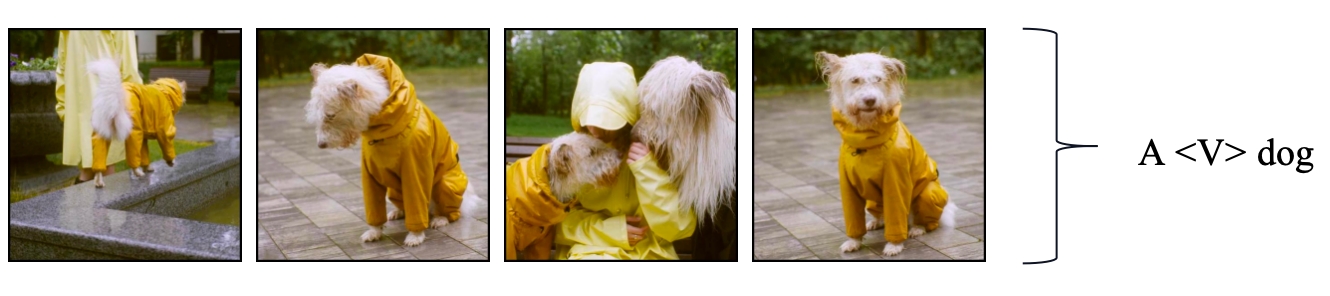}} \\
    \subfloat[Prompt: A <V> dog lies on the ground and then goes to eat from the bowl.]{\includegraphics[width=\linewidth]{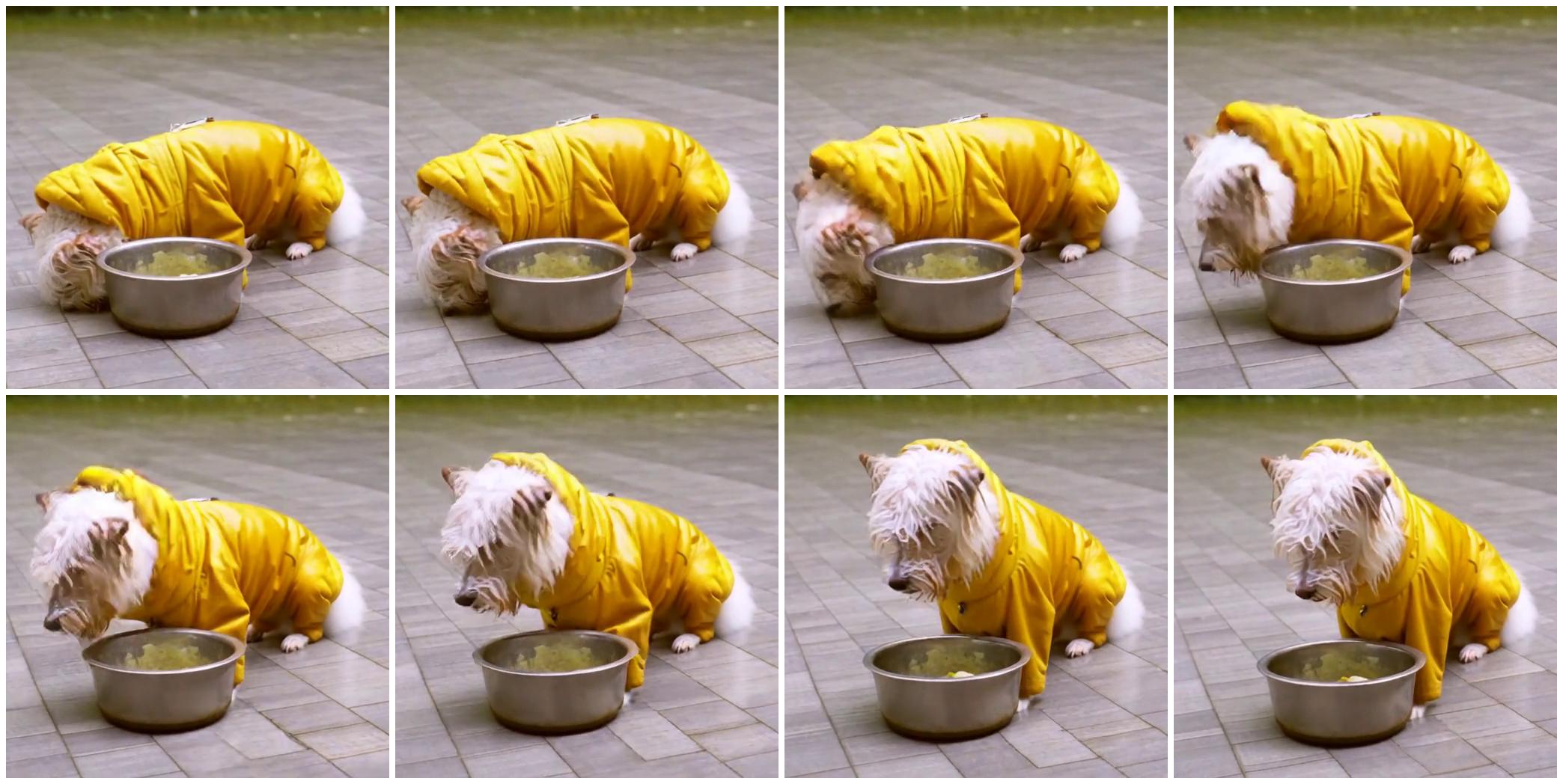}} \\
    \subfloat[Prompt: A <V> dog bit his tail happily and shakes his head.]{\includegraphics[width=\linewidth]{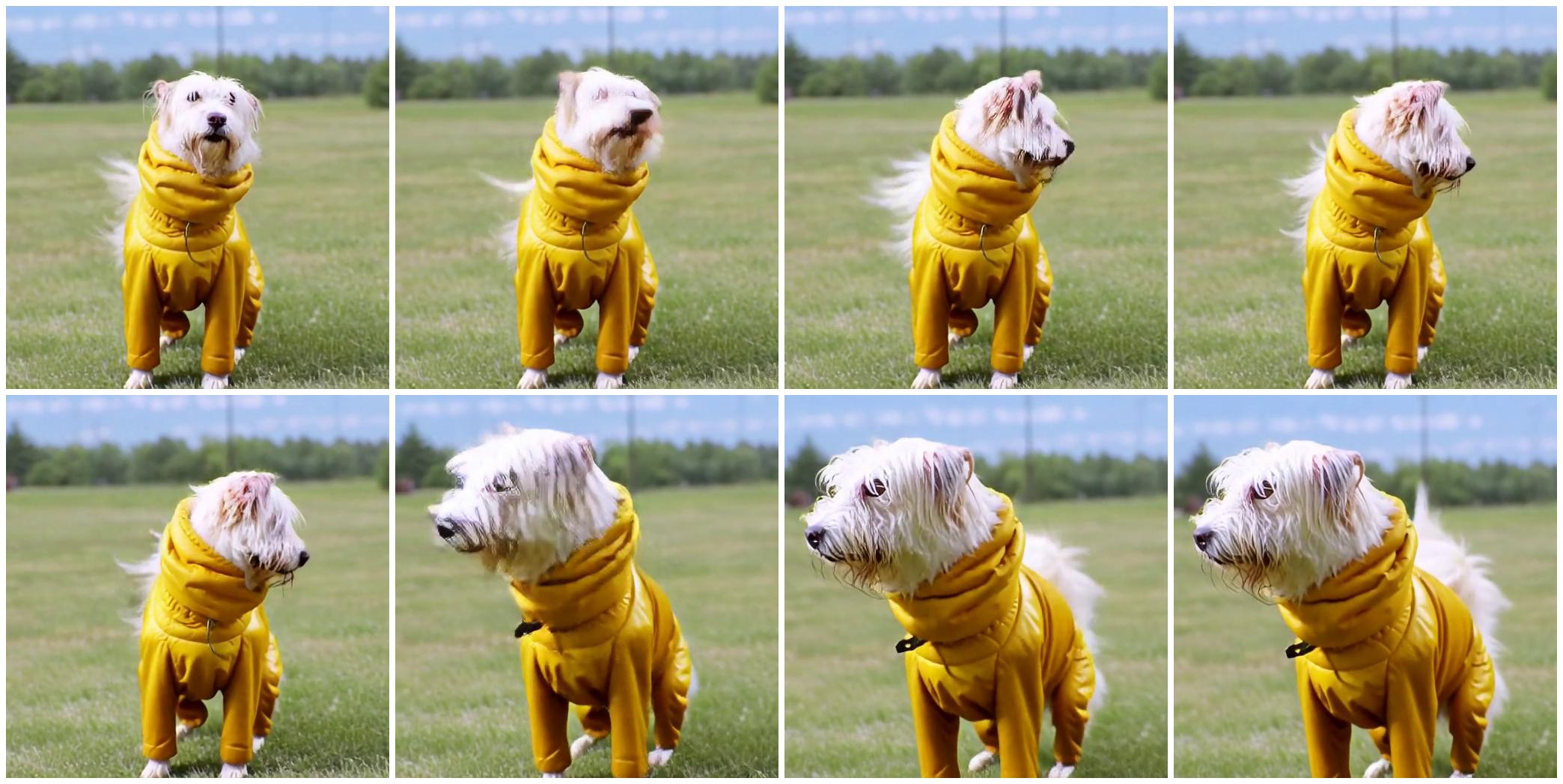}} \\
    \caption{Subject-driven generation examples of \name{}.}
    \label{fig:db}
\end{figure}

We surprisingly find that \name{} can perform subject-driven video generation by finetuning solely on images without videos. 
For example, we use the DreamBooth~\cite{ruiz2023dreambooth} technique to designate the learned subject as a special symbol <V> for finetuning. As shown in Figure~\ref{fig:db}, the generated videos faithfully recreates the learned subject.

\section{Conclusion}

We present \name{}, a high-definition text-to-video generator that demonstrates strong abilities in various aspects, including duration, coherence, and dynamism of the generated videos, on par with Sora. In the future, \name{} still has room for improvement. For instance, there are occasional flaws in details, and interactions between different subjects in the video sometimes deviate from physical laws. We believe that these issues can be effectively addressed by further scaling \name{}.

\section{Acknowledgements}

We appreciate the support of the data team and the product team for the project at Shengshu. This work was partly supported by NSFC Projects (Nos. 62061136001, 62106123, 61972224), Tsinghua Institute for Guo Qiang, and the High Performance Computing Center, Tsinghua University. J.Z is also supported by the
XPlorer Prize.

\bibliographystyle{plain}
\bibliography{refs}

\end{document}